\pdfoutput=1
\RequirePackage{fix-cm}
\documentclass[11pt]{article}

\usepackage[final]{acl}

\usepackage{times}
\usepackage{latexsym}

\usepackage[T1]{fontenc}

\usepackage[utf8]{inputenc}

\usepackage{microtype}

\usepackage{inconsolata}

\usepackage{times}
\usepackage{latexsym}
\usepackage{listings}
\usepackage{booktabs}
\usepackage{amsmath,amssymb}
\newtheorem{theorem}{Theorem}
\newtheorem{definition}{Definition}

\newtheorem{remark}{Remark}
\usepackage{float}
\usepackage{graphicx}
\usepackage{tikz}
\usepackage{subfigure}
\usepackage{graphicx}
\usepackage{mwe}
\usepackage{caption}
\usepackage{svg}
\usepackage{epstopdf}
\usepackage{wrapfig}
\usepackage{algorithm, algorithmic}
\usepackage{booktabs}
\usepackage{bm}
\usepackage{multirow}
\usepackage{colortbl, xcolor}
\usepackage{transparent}

\usetikzlibrary{arrows,automata}
\usetikzlibrary{positioning}
\definecolor{TopRow}{rgb}{0.13,0.67,0.8}
\definecolor{NormalRow}{rgb}{0.54,0.81,0.94}

\newcommand{\colorword}[3]{\colorbox{#1}{\textcolor{#2}{#3}}}

\newtheorem{Proof}{Proof}

\title{Private Language Models via Truncated Laplacian Mechanism}

\author{
  \textbf{Tianhao Huang\textsuperscript{1,2,3}\footnotemark[1]}, 
  \textbf{Tao Yang\textsuperscript{1,2,3}\footnotemark[1]}, 
  \textbf{Ivan Habernal\textsuperscript{4}},
  \textbf{Lijie Hu\textsuperscript{1,2}},
  \textbf{Di Wang\textsuperscript{1,2}\footnotemark[2]},
  \\
  \textsuperscript{1}Provable Responsible AI and Data Analytics (PRADA) Lab, \\
  \textsuperscript{2}King Abdullah University of Science and Technology, \textsuperscript{3}Nankai University, \\
  \textsuperscript{4}Research Center Trustworthy Data Science and Security of the University Alliance Ruhr,\\ Faculty of Computer Science, Ruhr University Bochum, \\
  \texttt{\{tianhao.huang, tao.yang, lijie.hu, di.wang\}@kaust.edu.sa} \\ \texttt{ivan.habernal@ruhr-uni-bochum.de} \\ 
}

\begin{document}
\maketitle

\renewcommand{\thefootnote}{\fnsymbol{footnote}}
\footnotetext[1]{Equal contribution. Part of the work was done as a research intern at PRADA Lab.}
\footnotetext[2]{Corresponding author.}

\begin{abstract}
Deep learning models for NLP tasks are prone to variants of privacy attacks.
To prevent privacy leakage, researchers have investigated word-level perturbations, relying on the formal guarantees of differential privacy (DP) in the embedding space. However, many existing approaches either achieve unsatisfactory performance in the high privacy regime when using the Laplacian or Gaussian mechanism, or resort to weaker relaxations of DP that are inferior to the canonical DP in terms of privacy strength. This raises the question of whether a new method for private word embedding can be designed to overcome these limitations.

In this paper, we propose a novel private embedding method called the high dimensional truncated Laplacian mechanism. Specifically, we introduce a non-trivial extension of the truncated Laplacian mechanism, which was previously only investigated in one-dimensional space cases. 
Theoretically, we show that our method has a lower variance compared to the previous private word embedding methods. To further validate its effectiveness, we conduct comprehensive experiments on private embedding and downstream tasks using three datasets. Remarkably, even in the high privacy regime, our approach only incurs a slight decrease in utility compared to the non-private scenario.
\end{abstract}

\section{Introduction}

\begin{figure*}
\centering
\tikzset{
    >=stealth',
    punkt/.style={
          very thick,
          rectangle split,
          rectangle split parts=2,
          inner ysep=1.2mm,
          inner xsep=1.5mm,
          rounded corners,
          draw=black, thick,
          text width=14.8cm},
    punx/.style={
          rectangle,
          rounded corners,
          align=left,
          text width=0.55cm,
          text depth=0.01cm,
          draw=white},
  header/.style = {%
    inner ysep = +1.5em,
    append after command = {
      \pgfextra{\let\TikZlastnode\tikzlastnode}
      node [header node] (header-\TikZlastnode) at (\TikZlastnode.north) {#1}
      node [span = (\TikZlastnode)(header-\TikZlastnode)]
        at (fit bounding box) (h-\TikZlastnode) {}
    }
  },
}

\begin{tikzpicture}[scale=1.0]
\node[punkt](Vanilla){
{\textbf{Comparison of Private Embedding} \quad }
\nodepart{second}
{\centering \small \textbf{Original}:~~~\,
\colorword{pink!20}{black}{Oh and we came on a Saturday night around} \colorword{pink!100}{black}{11:30} 
\colorword{pink!20}{black}{for} 
\colorword{pink!100}{black}{context.}
($\rightarrow$\textcolor{red}{Privacy Leakage})
}\\
{\centering \small \textbf{Trlaplace}:~~\, 
\colorword{pink!20}{black}{Oh and we came on a Saturday night around} \colorword{pink!100}{black}{9:30pm} 
\colorword{pink!20}{black}{for} 
\colorword{pink!100}{black}{<unk>}
($\rightarrow$\textcolor{blue}{Private and Fluent})
}\\
{\centering \small \textbf{Laplace}:~~~\,
\colorword{pink!20}{black}{Oh and we came on a Saturday night around} \colorword{pink!100}{black}{around} 
\colorword{pink!20}{black}{for} 
\colorword{pink!100}{black}{<unk>}
($\rightarrow$\textcolor{red}{Semantic Problem})
}\\
{\centering \small \textbf{Gaussian}:~~\,
\colorword{pink!20}{black}{Oh and we came on a Saturday night around} \colorword{pink!100}{black}{11:30} 
\colorword{pink!20}{black}{for} 
\colorword{pink!100}{black}{<unk>}
($\rightarrow$\textcolor{red}{Privacy Leakage})
}
};
\end{tikzpicture}
\caption{An example of (private) text re-write for different mechanisms with $\epsilon=0.1$. \label{fig:exp}}
\end{figure*}

The recent developments of deep learning have led to significant success in various tasks in Natural Language Processing (NLP), from next word prediction
in mobile keyboards \citep{DBLP:journals/corr/abs-1906-04329}, to critical tasks like
predicting patient health conditions from clinical records
\citep{DBLP:journals/midm/YaoML19}. However, such applications may always involve user-generated textual data as the training dataset, which contains sensitive information. To address privacy concerns, text anonymization \citep{anandan2012t,pilan2022text} has been commonly used, which involves identifying sensitive attributes and replacing them with alternative values. Nevertheless, such heuristic approaches become ineffective as deep neural networks often tend to memorize training data, making them susceptible to information leakage about the training data \citep{shokri2017membership,carlini2021extracting,carlini2019secret}. One way that takes into account the limitations of existing approaches is designing Differentially Private (DP) algorithms. Differential privacy~\citep{dwork2006our} is resilient to arbitrary side information that might be available to attackers and has become a de facto method for private data analysis~\citep{xiang2024preserving,DBLP:journals/jmlr/0015GS020,DBLP:journals/pacmmod/Xiang0L023,wang2020differentially,su2022faster,wang2019estimating,hu2023privacy,wang2023generalized,wang2017differentially,wang2019differentially,wang2018empirical,DBLP:conf/ijcai/XueYH021,DBLP:conf/ijcai/HuaiWM0Z19,DBLP:conf/ecai/0015DHXP023,DBLP:journals/tcs/WangX20,DBLP:conf/aaai/HuaiWM0Z20,DBLP:conf/pods/HuNX022}. 

Recently, there has been significant research focusing on differentially private (DP) versions of word embedding from various perspectives \citep{yue-etal-2021-differential,feyisetan2019leveraging,krishna-etal-2021-adept,feyisetan2020privacy,xu2021density,xu2021utilitarian,carvalho2021tem,DBLP:journals/corr/abs-2107-07923,habernal-2021-differential,DBLP:conf/acl/Habernal22}. However, there are still some shortcomings in these approaches. On the one hand, several works consider adding Laplacian or Gaussian noise to the embedding space to ensure DP \cite{habernal-2021-differential,krishna-etal-2021-adept,DBLP:conf/acl/Habernal22}.  
However, these mechanisms suffer from high noise levels, resulting in low utility, especially in the high privacy regime when the privacy parameter ($\epsilon$) is small. Moreover, these mechanisms can even alter the semantics of sentences (see Fig.\ref{fig:exp}). 
On the other hand, there is a growing body of work that focuses on a relaxation of the canonical definition of DP, known as metric DP, which can achieve better performance. However, as a relaxed notion of DP, Metric DP cannot provide the same level of strong privacy guarantees as the canonical DP \citep{mattern-etal-2022-limits}. This raises the question of \textit{whether we can develop improved private word embedding mechanisms within the framework of canonical DP that can have comparable performance with existing metric DP-based methods}.

In this paper, we provide an affirmative answer to the previous question by proposing a novel private mechanism for word embedding. Our approach is inspired by the superior performance of the truncated Laplacian mechanism in one-dimensional space \cite{DBLP:conf/aistats/GengDGK20}. However, it remains unclear whether this superiority can extend to high dimensional cases, as directly extending the one-dimensional truncated Laplacian mechanism is challenging. To bridge this gap, we develop a high dimensional truncated Laplacian mechanism (TrLaplace), which is a non-trivial extension of the one-dimensional case. Theoretically, we show that compared with Laplacian and Gaussian mechanisms for private word embedding, TrLaplace-based private embedding has a lower variance. Moreover, we also conduct intensive experiments on both private embedding and downstream tasks to show our approach significantly outperforms the previous DP-based methods in the high privacy regime, and it will not drop much accuracy and utility compared with the non-private case. Moreover, compared to the existing metric DP-based method, our mechanism has even better performance for privacy tests while also keeping comparable performance for downstream tasks. 

\section{Related Work}
Recent years have seen substantial advancements in language models within differential privacy (DP) frameworks. Due to the space limit, here we only mention the existing literature on private word embedding. We refer the readers to the survey \citep{hu-etal-2024-differentially} for more details. 

Current research on private word embeddings can be broadly categorized into two approaches: original DP-based methods and metric DP-based methods. The seminal work in the original DP category by \citet{lyu2020towards} introduces a framework utilizing the Unary Encoding mechanism. This approach was subsequently refined by \citet{plant-etal-2021-cape}. Further improvements were made by \citet{lyu-etal-2020-differentially}, who proposed a dropout technique for perturbed embeddings to enhance downstream task fairness. However, \citet{qu2021natural} identify a critical privacy issue in \citet{lyu-etal-2020-differentially}, noting that it requires access to users' raw data for fine-tuning during the training phase. Other notable contributions include works by \citet{krishna-etal-2021-adept}, \citet{habernal-2021-differential}, and \citet{DBLP:conf/sbp-brims/AlnasserB021}, who explore privatizing word embeddings. \citet{krishna-etal-2021-adept} and \citet{DBLP:conf/sbp-brims/AlnasserB021} propose ADePT, an auto-encoder-based DP algorithm. Unfortunately, \citet{habernal-2021-differential} points out that ADePT is not differentially private by thorough theoretical proof. \citet{Igamberdiev.2022.COLING} address reproducibility by providing source code for DP Auto-Encoder methods. In this paper, we aim to improve the performance of the mechanisms in \citet{Igamberdiev.2022.COLING}.

In the realm of metric DP, \citet{feyisetan2020privacy} first study this problem and provide a general perturbation-and-projection framework. \citet{xu2020mahala} reconsider this problem setting, replacing the Euclidean distance with the Mahalanobis distance to improve the utility. Subsequently, \citet{xu2021utilitarian} introduce the Vickrey mechanism to further refine the utility in the projection step. To address the limitations of the multivariate Laplace mechanism, \citet{xu2021density} and \citet{carvalho2021tem} propose a Truncated Gumbel Noise method. \citet{feyisetan2021private} tackle high-dimensionality issues using random projection. Additionally, \citet{feyisetan2019leveraging} define hyperbolic embeddings and utilize the Metropolis-Hastings algorithm for sampling from hyperbolic distributions. More recently, \citet{tang2020privacy} explore differential privacy with varying privacy levels for different words. \citet{DBLP:journals/corr/abs-2306-01457} introduce sense embeddings with a sense disambiguation step prior to noise injection, and \citet{DBLP:journals/corr/abs-2306-01471} address common semantic context issues in prior private embedding mechanisms. It is crucial to clarify that the objective of this work is the privatization of embedded outputs rather than the embedded methods themselves. The pre-trained initial embedding methods and the corresponding embeddings of all words in the vocabulary are treated as public knowledge. This distinction is significant because it allows us to perform projections without incurring additional privacy costs. By leveraging publicly available pre-trained embeddings, our method demonstrates how effective privacy-preserving techniques can be implemented while still utilizing existing resources.

\section{Preliminaries}

Differential Privacy is a data post-processing technique designed to ensure data privacy by adding confusion to potential attackers. Specifically, suppose there is one dataset noted as $\mathcal{D}$, and we change or delete one data record in this dataset which we call $\mathcal{D'}$. If the output distributions of $\mathcal{D}$ and $\mathcal{D'}$ are close enough, then we cannot distinguish these two distributions, i.e., we cannot infer whether the deleted or replaced data sample is really in this dataset. The formal details are given by \citep{dwork2006our}. Note that in the definition of DP, adjacency is a key notion. 
One of the commonly used adjacency definitions
is that two datasets $S$ and $S'$ are adjacent (denoted as $S\sim S'$) if $S'$
can be obtained by modifying one record in $S$.

\begin{definition}\label{def:dp}
Given a domain of dataset $\mathcal{X}$.
A randomized algorithm $\mathcal{A}: \mathcal{X}\mapsto \mathcal{R}$ is $(\varepsilon,\delta)$-differentially private (DP) if for all adjacent datasets $S,S'$ with each sample is in $\mathcal{X}$ and for all $T\subseteq \mathcal{R}$, the following holds
	$$\Pr(\mathcal{A}(S) \in T)\leq \exp(\varepsilon) \Pr(\mathcal{A}(S')\in T)+\delta.$$
	When $\delta=0$, we call the algorithm $\mathcal{A}$ is $\varepsilon$-DP. 
\end{definition}

In this work, we adopt a similar setting to previous research on private word embedding \citep{feyisetan2020privacy,xu2021density,krishna-etal-2021-adept}. We consider a scenario where a user inputs a word $w$ from a discrete fixed vocabulary $\mathcal{W}$. Our goal is to preserve the user's privacy with respect to her/his word. To achieve this goal, we aim to design an algorithm that accepts $w$ as input and whose distribution of output is close to the case where $w'\in \mathcal{W}$ is the input, with $w'\neq w$ is any other word. From the attacker's perspective, based on the output, he cannot distinguish whether the user's input word is $w$ or $w'$ as their output distributions are almost the same. Formally, we have the following definition.

\begin{definition}\label{def:1}
	Given a discrete vocabulary  $\mathcal{W}$, a randomized algorithm $\mathcal{A}: \mathcal{W}\mapsto \mathcal{R}$ is word-level $(\epsilon,\delta)$-differentially private (DP) if for all pair of words $w, w'\in \mathcal{W}$ and for all $T\subseteq \mathcal{R}$ we have 
	$\mathbb{P}(\mathcal{A}(w)\in T)\leq e^{\epsilon} \mathbb{P}(\mathcal{A}(w')\in T)+\delta.$
	When $\delta=0$, we call the algorithm $\mathcal{A}$ is $\epsilon$-DP. 
\end{definition}
In this paper, we assume the user holds a sentence $s=w_1w_2\cdots w_n$ with $n$ words. And we aim to design an $(\epsilon, \delta)$-DP algorithm, which is private w.r.t. each word $w_i$.

\section{Private Embedding via Truncated Laplacian Mechanism}

In this section, we will provide details of our method. Generally speaking, for each token $w_i$, to achieve DP, our approach consists of three steps. First, each token $w_i$
is mapped to an
$d$-dimensional pre-trained word embedding $\phi(w_i)$.  And we perform a clipping step to get a clipped embedding: 
\begin{equation}\label{eq:1}\small
\operatorname{CLIPEmb}(w_i)=\phi(w_i)\min\{1, \frac{C}{\|\phi(w_i)\|_2}\}, 
\end{equation}
where the threshold $C>0$ is a hyper-parameter. In the second step, we add some random noise to the clipped embedding vector to make it satisfies DP. Finally, we will perform the projection step by finding the nearest word $\hat{w}_i$ to the perturbed and clipped embedding vector within the embedding space: 

\begin{equation}\label{eq:em}
   \hat{w}_i=\arg\min_{w\in\mathcal{W}}\|\phi(w)- \operatorname{CLIPEmb}(w_i)-\eta\|_2, 
\end{equation}
where $\eta$ is the randomized noise we add in the second step. See Algorithm \ref{alg:1} for details. 
\begin{algorithm}
	\renewcommand{\algorithmicrequire}{\textbf{Input:}}
	\renewcommand{\algorithmicensure}{\textbf{Output:}}
	\caption{Privacy Preserving Mechanism\label{alg:1}}
	\begin{algorithmic}[1]
	\REQUIRE String $s=w_{1} w_{2} \ldots w_{n}$, clipping threshold $C$, privacy parameter $\epsilon>0$.
	\ENSURE String $\hat{s}=\hat{w}_{1}\hat{w}_{2} \ldots \hat{w}_{n}$.
	
	\FORALL{$i \in\{1, \ldots, n\}$}
	    \STATE Sample $\mathbf{\eta}$ from the truncated Laplacian distribution in Theorem \ref{thm:1}.
	    \STATE Obtain the perturbed clipped embedding $\mathbf{r}_i= \operatorname{CLIPEmb}(w_i)+\mathbf{\eta}$.
	    \STATE
	    Let  $\hat{w}_i=\operatorname{Proj}\left(\mathbf{r_i}\right)$ as in (\ref{eq:em}). 
	\ENDFOR
    \STATE \textbf{return} 
    $\hat{s}=\hat{w}_{1}\hat{w}_{2} \ldots \hat{w}_{n}$.
	\end{algorithmic}  
\end{algorithm}

It is notable that the goal of clipping is to make the $\ell_2$-norm of embedding vector be bounded  so that we can
adding noise to ensure DP, such as the Laplacian mechanism or Gaussian mechanism \cite{Dwork2014TheAF}.

\begin{theorem}[Laplacian Mechanism]\label{lap}
Suppose $\operatorname{CLIPEmb}(\mathbf{w}) \in \mathbb{R}^d$ denote the clipped embedding vector with threshold $C$. Then the mechanism $\mathcal{A}_{lap}(w)=\operatorname{CLIPEmb}(w)+\eta_1$ is $\epsilon$-DP, where $\eta_1=(\eta_{1,1},\cdots, \eta_{1, d})$ and $\eta_{i,j}$ is drawn from a Laplacian Distribution $\text{Lap}(\frac{\Delta_1(f)}{\epsilon})$ with $\Delta_1=2\sqrt{d}C$. For a parameter $\lambda$, the Laplacian distribution has the density function $\text{Lap}(\lambda) (x)=\frac{1}{2\lambda}\exp(-\frac{x}{\lambda})$. 
\end{theorem}

\begin{theorem}[Gaussian Mechanism]\label{gau}
Suppose $\operatorname{CLIPEmb}(\mathbf{w}) \in \mathbb{R}^d$ denote the clipped embedding vector with threshold $C$. Then the mechanism $\mathcal{A}_{gau}(w)=\operatorname{CLIPEmb}(w)+\eta_2$ is $(\epsilon, \delta)$-DP when $\epsilon\leq 1$, where $\eta_2\sim \mathcal{N}(0, \frac{8C^2\ln(1.25/\delta)}{\epsilon^2}I_d)$ is drawn from a Gaussian distribution. 
\end{theorem}

In the following we propose an improved mechanism namely high dimensional truncated Laplacian mechanism. Before that we first recall the probability density function of the one-dimensional truncated Laplacian distribution, which could be written as the following with some appropriate constants $\alpha, A$ and $B$:
\begin{equation}\label{eq:2}
f_{TLap}(x)=\begin{cases}\frac{1}{B} e^{-\alpha|x|}, & \text { for } x \in[-A, A] \\ 0, & \text { otherwise. }\end{cases} 
\end{equation}
In our mechanism, we add  high dimensional truncated Laplacian noise to the clipped embedding vector. Here each coordinate of the noise  is i.i.d. sampled from a truncated Laplacian distribution with some specific $\alpha, A$ and $B$. 

\begin{remark}
It is notable that although using truncated Laplacian noise to ensure DP has been studied quite well \cite{DBLP:conf/aistats/GengDGK20,DBLP:journals/corr/abs-2107-12957}, all of them only considered the case where the dimension $d=1$ and their methods cannot extend to the case where $d>1$. For example, \cite{DBLP:conf/aistats/GengDGK20} only shows that adding noise with density function (\ref{eq:2}) with $A=\frac{\Delta_1}{\epsilon}\log (1+\frac{e^\epsilon}{2\delta})$ and $\alpha=\frac{\epsilon}{\Delta_1}$ can ensure $(\epsilon, \delta)$-DP. Compared with the high dimensional case in Theorem \ref{thm:1} we can see the constant $A$ is more complicated and the proof is also different. Thus, our mechanism cannot be considered as a trivial extension of the one-dimensional truncated Laplacian mechanism. Secondly, while the Laplacian mechanism can guarantee $\epsilon$-DP, the truncated one can only ensure $(\epsilon, \delta)$-DP, which is the same as in the one-dimension case. 
\end{remark} 

However, as we will see below, our mechanism is superior to Laplacian  mechanism for utility. It is also notable that we need to assume $\epsilon \leq 2\delta^\frac{1}{d}\sqrt{d}$, this is reasonable since we always wish $\epsilon$ to be as small as possible, as large $\epsilon$ indicates the algorithm is no longer private. If we want large $\epsilon>2\delta^\frac{1}{d}\sqrt{d}$, we can use the trick of adding dummy dimension to the vector to increase its dimensionality manually and then projecting back to the original space after adding noise.

\begin{figure}[t]
    \centering
    \includegraphics[width=0.75\linewidth]{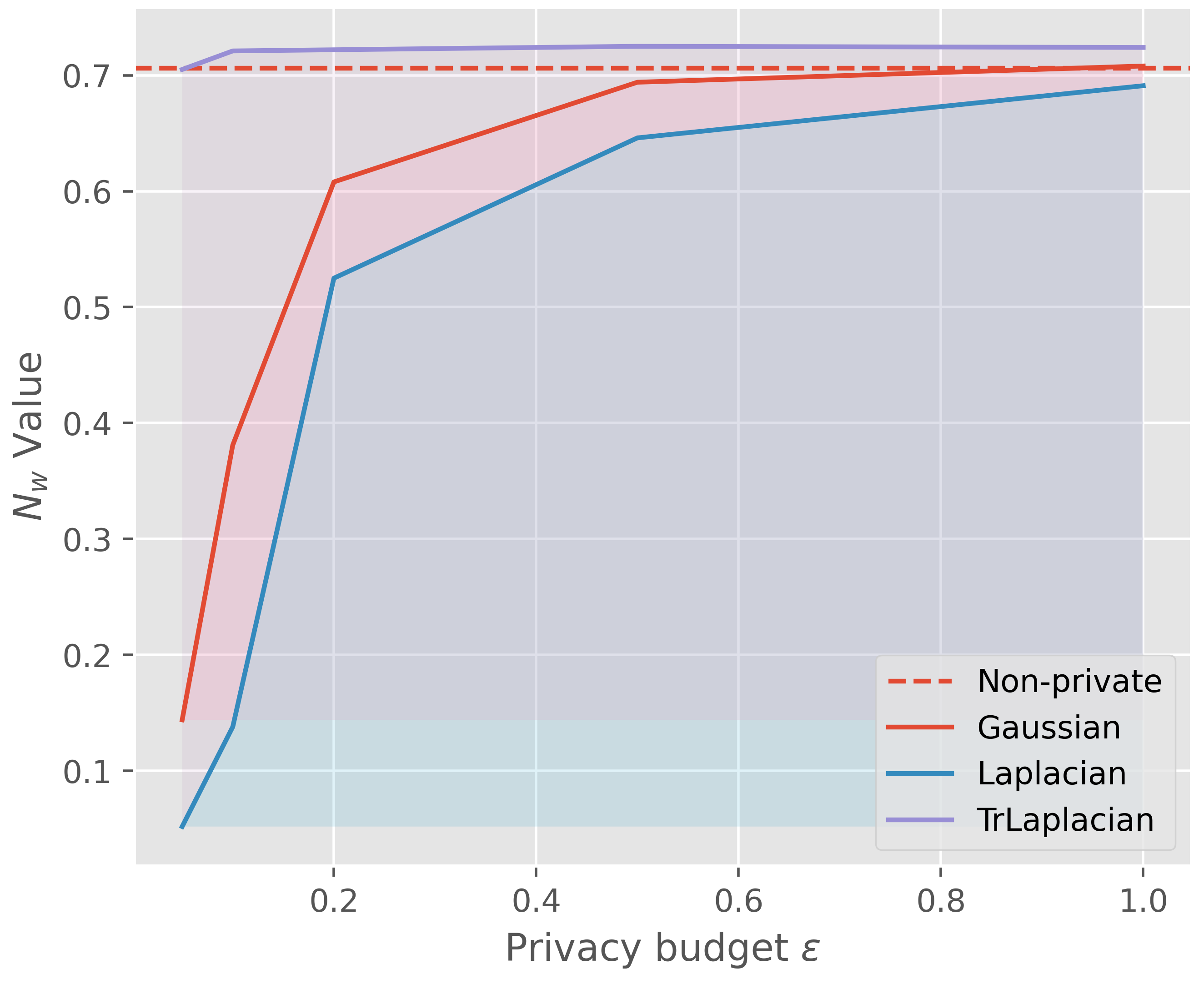}
    \caption{Privacy Test.  Curves of the value $N_w$  with privacy budget $\epsilon$ for Yelp dataset.\label{fig:nw}}
\end{figure}

\section{Theoretical Sensitivity Analysis}
In the last section, we introduce our truncated laplacian mechanism, we will analyze its sensitivity and proof our claim in this section.

\begin{theorem}\label{thm:1}
Suppose $\operatorname{CLIPEmb}(w) \in \mathbb{R}^d$ is the clipped embedding vector with threshold $C$. Define $\Delta_{\infty}=2C$ and $\Delta_{1}=2\sqrt{d}C$. For $\epsilon \leq 2\delta^\frac{1}{d}\sqrt{d}$, if 
\begin{align*}\small
&\alpha=\frac{\epsilon}{\Delta_{1}}, A= -\frac{\Delta_{1}}{\epsilon} \log (1-\frac{\epsilon}{2\delta^\frac{1}{d}\sqrt{d}})\\
& B=\frac{2(1-e^{-\alpha  A})}{\alpha}=\frac{\Delta_\infty}{\delta^\frac{1}{d}}, 
\end{align*}
then the mechanism $\mathcal{A}(w)=\operatorname{CLIPEmb}(w)+\eta$ is $\left(\epsilon, \delta\right)$-DP, where $\eta=(\eta_1, \cdots, \eta_1)$ and each $\eta_i$ has the density function as in (\ref{eq:2}) with the above parameters. 
\end{theorem}

\begin{table*}[t]
\caption{\textbf{Privacy Test.} Performance under fastText Embedding initialization for the non-private case ($\epsilon=\infty$) and  three mechanisms (Gaussian, Laplacian and TrLaplacian) on Yelp dataset. The privacy budget ranges from 0.05 to 20. $\uparrow$ means a higher value under this metric indicates better results, and $\downarrow$ means the opposite. The best performance is \textbf{bolded}. The same symbols are used in the following tables by default.\label{tb:6}}
\resizebox{\textwidth}{!}{%
\begin{tabular}{cccccccccccccccccc}
\toprule 
&  & \textbf{Original} &  & \multicolumn{4}{c}{\textbf{Gaussian}} &  & \multicolumn{4}{c}{\textbf{Laplacian}} &  & \multicolumn{4}{c}{\textbf{TrLaplacian}} \\ \cmidrule{3-3} \cmidrule{5-8} \cmidrule{10-13} \cmidrule{15-18} 
\textbf{Privacy budget \bm{$\epsilon$}} &  & $\infty$ &  & 0.05     & 0.1    & 0.2    & 0.5    &  & 0.05     & 0.1    & 0.2    & 0.5    &  & 0.05              & 0.1    & 0.2    & 0.5    \\ 
\midrule

\textbf{Loss\bm{$\downarrow$}}             &  &   3.35  & & 35.01  &29.33   & 9.31  & 4.50  & &36.23  & 29.69  & 17.15  &  5.58 &  & \textbf{1.20}  & \textbf{1.20}  & \textbf{1.26}   & \textbf{1.23}  \\
 \rowcolor{gray!30}
\textbf{Rouge1\bm{$\uparrow$}}           &  &  87.8 &  & 12.72 & 28.68 & 77.95  & 86.90 &  & 10.99 & 27.96 & 58.97 & 85.16 &  & \textbf{92.43} & \textbf{92.67}  & \textbf{92.29} & \textbf{92.43} \\

\textbf{BLEU\bm{$\uparrow$}}             &  &  8.929  &  & 8.226 & 8.745  & 8.918 & 8.931 &  & \textbf{8.998} & 8.681   & 8.898 & 8.931 &  & 8.937  & \textbf{8.938} & \textbf{8.937} & \textbf{8.938} \\

\rowcolor{gray!30}
\bm{$N_w\uparrow$}     &  & 0.713 &  & 0.138  & 0.232 & 0.661 & 0.765 &  & 0.058 & 0.225 &  0.484 & 0.753 & &  \textbf{0.813} & \textbf{0.807} & \textbf{0.804} & \textbf{0.813} \\

\textbf{BERT-S\bm{$\uparrow$}}             &  & 0.967 &  & 0.864  & 0.873 & 0.945 & 0.966 &  & 0.857 & 0.867 & 0.908  & 0.962 &  &  \textbf{0.981}  & \textbf{0.978} & \textbf{0.979} & \textbf{0.978} \\
\midrule
\midrule

&  & \textbf{Original} &  & \multicolumn{4}{c}{\textbf{Gaussian}} &  & \multicolumn{4}{c}{\textbf{Laplacian}} &  & \multicolumn{4}{c}{\textbf{TrLaplacian}} \\ \cmidrule{3-3} \cmidrule{5-8} \cmidrule{10-13} \cmidrule{15-18} 
\textbf{Privacy budget \bm{$\epsilon$}} &  & $\infty$ &  & 1     & 5    & 10    & 20    &  & 1     & 5    & 10    & 20    &  & 1              & 5    & 10    & 20    \\
\midrule
 
\textbf{Loss\bm{$\downarrow$}}             &  &  
3.35 &  & 3.10  & 1.68   & 1.48  & 1.29   &  & 3.60  & 1.55  &  1.53  & 1.51  &  & \textbf{1.22}  & \textbf{1.25}  & \textbf{1.28}  & \textbf{1.27} \\

\rowcolor{gray!30}
\textbf{Rouge1\bm{$\uparrow$}}           &  &   87.8  &  & 89.47&   92.06 & \textbf{92.40} & \textbf{92.49} & & 88.17 & 91.87 & 91.90 & 91.91&  & \textbf{92.42} & \textbf{92.35}  & 92.34  & 92.31 \\

\textbf{BLEU\bm{$\uparrow$}}        &  & 8.929 &  &  8.936 & 8.937 & 8.936 & 8.936 &  & 8.935 & 8.937 & 8.936  & 8.934 &  & \textbf{8.938}  & \textbf{8.939} & \textbf{8.937} & \textbf{8.938} \\

\rowcolor{gray!30}
\bm{$N_w\uparrow$}     &  & 0.713 &  &  0.794 & \textbf{0.809} & \textbf{0.804} & \textbf{0.813} &  & 0.758 & 0.801 & 0.795  & 0.792  &  & \textbf{0.807}  & 0.802 & 0.800 &0.808 \\

\textbf{BERT-S\bm{$\uparrow$}}     &  & 0.967 &  & 0.976     & 0.977 & \textbf{0.978} & \textbf{0.980} &  & 0.967 & \textbf{0.978} & 0.976  & 0.977  &  & \textbf{0.979} & \textbf{0.978} & \textbf{0.978} & \textbf{0.980} \\
\bottomrule
\end{tabular}%
}
\end{table*}

In the following, we will show our mechanism has lower variance than the Laplacian and Gaussian mechanism, which indicates that our method is superior theoretically. 

\begin{theorem}\label{thm:2}
The variance of mechanism $\mathcal{A}$ in Theorem \ref{thm:1} is lower than the variance of Laplacian mechanism  and Gaussian mechanism when $\delta\leq \frac{1}{e^d}$. 
\end{theorem}

\section{Experiments}
In this section, we conduct experiments for our method based on two parts: DP text re-write for fine-tuning (private embedding) and downstream tasks (sentiment analysis). We have open-sourced our code on \url{https://github.com/kaustpradalab/TrLap}.

\subsection{Experimental Setup}
\paragraph{Datasets.}
For the DP text re-write task, we use the Yelp \footnote{https://www.yelp.com/dataset/} and Yahoo \citep{yang-etal-2019-read} datasets. The Yelp Open dataset is a subset of Yelp business, review, and user data with a training size of 8,539 and a testing size of 2,174. The Yahoo dataset contains 14,180 news articles and 34,022 click events. All data are collated to obtain a training, validation, and testing set segmented by sentences. 

For downstream tasks, we use the SST-2 dataset \citep{socher-etal-2013-recursive} for the sentiment analysis task, from which we use 68,221 heavily polarized reviews from the Internet Movie Database. We divide the SST-2 dataset into an 80:20 ratio for training and testing. The training set consists of 54,576 reviews and the testing set consists of 13,645 reviews. We use the AG News dataset \citep{zhang2015character} which includes news articles on the four main topics in the AG News corpus for the topic classification task. Following \citet{meisenbacher2024comparative}, we randomly select a sample of 6,000 articles from each topic, totaling 16,000 for training, 4,000 for validation, and 4,000 for testing. The statistics of the datasets are presented in Tab. \ref{tb:7} in Appendix \ref{Appendix1: datastatistics}.


\begin{table*}[!ht]
\caption{\textbf{Privacy Test.} Performance under GloVe Embedding initialization for the non-private case (  $\epsilon=\infty$) and  the three mechanisms, where the privacy budget ranges from 0.05 to 0.5. \label{tb:1}}
\resizebox{\textwidth}{!}{%
\begin{tabular}{ccccccccccccccccccc}
\toprule 
 & &  & \textbf{Original} &  & \multicolumn{4}{c}{\textbf{Gaussian}} &  & \multicolumn{4}{c}{\textbf{Laplacian}} &  & \multicolumn{4}{c}{\textbf{TrLaplacian}} \\ \cmidrule{4-4} \cmidrule{6-9} \cmidrule{11-14} \cmidrule{16-19} 
\multicolumn{2}{c}{\textbf{Privacy budget $\bm{\epsilon}$}} &  & $\infty$ &  & 0.05     & 0.1    & 0.2    & 0.5    &  & 0.05     & 0.1    & 0.2    & 0.5    &  & 0.05              & 0.1    & 0.2    & 0.5    \\ 

\midrule

\rowcolor{gray!30}
& \textbf{Loss\bm{$\downarrow$}}  &  &   2.95  &  & 51.25  & 26.66  & 9.92  & 5.97  &  & 51.43  & 37.86  & 15.35  &  7.31 &  & \textbf{2.89}  &  \textbf{2.86} & \textbf{2.84}  & \textbf{3.04}  \\

& \textbf{Rouge1\bm{$\uparrow$}}     &  &    92.37  &  & 14.01 & 59.52 & 83.61  & 89.06 &  & 13.02 & 43.30 & 75.77 & 86.98 &  & \textbf{92.44} & \textbf{92.43}  & \textbf{92.41} & \textbf{92.25} \\

\rowcolor{gray!30}
\textbf{Yahoo} & \textbf{BLEU\bm{$\uparrow$}}     &  &  8.501  &  & \textbf{9.286} & 8.418  & 8.489 & 8.499 &  & 9.132 & 8.287   & 8.474 & 8.493 &  & 8.499  & \textbf{8.500} & \textbf{8.497} & \textbf{8.504} \\

& \bm{$N_w\uparrow$}    &  & 0.703 &  & 0.072  & 0.511 & 0.595 & 0.628 &  & 0.066 & 0.334 &  0.566 & 0.642 & &  \textbf{0.706} & \textbf{0.682} & \textbf{0.666} & \textbf{0.662} \\

\rowcolor{gray!30}
& \textbf{BERT-S\bm{$\uparrow$}}     &  & 0.975 &  & 0.849  & 0.908 & 0.955 & 0.963 &  & 0.839 & 0.889 & 0.942  & 0.959 &  &  \textbf{0.976}  & \textbf{0.971} & \textbf{0.971} & \textbf{0.971} \\

\midrule

& \textbf{Loss\bm{$\downarrow$}}     &  & 3.07 &  & 34.67  & 21.62   & 10.61  & 5.98  &  & 36.00  & 34.64  &  14.86  & 7.38  &  & \textbf{2.98}  & \textbf{2.99}  & \textbf{3.02}  &  \textbf{2.94} \\

\rowcolor{gray!30}
& \textbf{Rouge1\bm{$\uparrow$}}  &  &   89.40  &  & 15.97 & 48.89 & 76.48 & 84.97 &  & 12.60 & 14.68 & 66.62 & 82.08 &  & \textbf{89.45} & \textbf{89.47}  & \textbf{89.34}  & \textbf{89.54} \\

\textbf{Yelp} & \textbf{BLEU\bm{$\uparrow$}}  &  & 8.934 &  & \textbf{8.976} & 8.850 & 8.926 & 8.930 &  & 8.607 & 8.916 & 8.913  & 8.928 &  &  8.931  & \textbf{8.935} & \textbf{8.936} & \textbf{8.936} \\

\rowcolor{gray!30}
& \bm{$N_w\uparrow$}   &  & 0.706 &  &  0.144 & 0.381 & 0.608 & 0.694 &  & 0.052 & 0.138 & 0.525  & 0.646 &  & \textbf{0.705}  & \textbf{0.721} & \textbf{0.722} & \textbf{0.725} \\
& \textbf{BERT-S\bm{$\uparrow$}}  &  & 0.973 &  & 0.874     & 0.895 & 0.943 & 0.964 &  & 0.855 & 0.874 & 0.927  &0.952  &  & \textbf{0.971}  & \textbf{0.973} & \textbf{0.971} & \textbf{0.972} \\ \bottomrule
\end{tabular}%
}
\end{table*}

\begin{table*}[!t]
\caption{\textbf{Privacy Test.} Performance under GloVe Embedding initialization for the non-private case (  $\epsilon=\infty$) and  the three mechanisms, where the privacy budget ranges from 1 to 20. \label{tb:3}}
\resizebox{\textwidth}{!}{%
\begin{tabular}{ccccccccccccccccccc}
\toprule 
 & &  & \textbf{Original} &  & \multicolumn{4}{c}{\textbf{Gaussian}} &  & \multicolumn{4}{c}{\textbf{Laplacian}} &  & \multicolumn{4}{c}{\textbf{TrLaplacian}} \\ \cmidrule{4-4} \cmidrule{6-9} \cmidrule{11-14} \cmidrule{16-19} 
\multicolumn{2}{c}{\textbf{Privacy budget $\bm{\epsilon}$}} &  & $\infty$ &  & 1     & 5    & 10    & 20    &  & 1     & 5    & 10    & 20    &  & 1              & 5    & 10    & 20    \\ 
\midrule

\rowcolor{gray!30}
& \textbf{Loss\bm{$\downarrow$}}   &  &   2.95  &  & 4.28  & 3.01  & 3.03  & 2.98  &  & 4.93  & 3.24  & 3.05  & 3.13  &  & \textbf{2.85}  & \textbf{2.97}  &  \textbf{2.92} & \textbf{2.81}  \\

& \textbf{Rouge1\bm{$\uparrow$}}  &  &    92.37  &  & 90.97 & 92.27 & 92.16  & 92.19 &  & 90.02 & 92.09 & \textbf{92.28} & 92.26 &  & \textbf{92.41} &  \textbf{92.35} & 92.24 & \textbf{92.45} \\

\rowcolor{gray!30}
\textbf{Yahoo} & \textbf{BLEU\bm{$\uparrow$}}   &  &  8.501  &  & 8.501 & 8.501  & 8.499 & \textbf{8.500} &  & \textbf{8.503} &  8.501  & \textbf{8.502} & 8.500 &  & 8.498  & \textbf{8.501} & 8.499 & 8.499 \\

& \bm{$N_w\uparrow$}  &  & 0.703 &  &  0.637 & 0.680 & 0.664 & 0.672 &  & 0.660 & 0.658 & 0.675  & 0.655 &  &  \textbf{0.674}  & \textbf{0.670} & \textbf{0.702} & \textbf{0.680} \\

\rowcolor{gray!30}
& \textbf{BERT-S\bm{$\uparrow$}}  &  & 0.975 &  & 0.968     & \textbf{0.973} & 0.971 & 0.972 &  & 0.966 & 0.970 & 0.971  &0.971  &  & \textbf{0.974}  & 0.972 & \textbf{0.975} & \textbf{0.974} \\

\midrule

\multirow{5}{*}{\textbf{Yelp}} & \textbf{Loss\bm{$\downarrow$}}     &  &  3.07    &  &  4.74 & 3.14  & 3.13  & 2.97  &  & 5.02  & 3.30  & 3.66   & 3.17  &  & \textbf{2.93}  & \textbf{3.03}  & \textbf{3.00}  & \textbf{2.98}  \\

\rowcolor{gray!30}
& \textbf{Rouge1\bm{$\uparrow$}}      &  &   89.40  &  & 86.63 & 89.13 & 89.27 & \textbf{89.80} &  & 86.43 & 89.04 & 88.15 & 89.23 &  & \textbf{89.68} & \textbf{89.40}  & \textbf{89.37} & 89.60 \\

& \textbf{BLEU\bm{$\uparrow$}}     &  & 8.934 &  &  8.933 & \textbf{8.936} & 8.933 & \textbf{8.944} &  & 8.931 & 8.932 & 8.933  & 8.934 &  &  \textbf{8.934}  & 8.931 & \textbf{8.934} & 8.938 \\

\rowcolor{gray!30}
& \bm{$N_w\uparrow$}     &  & 0.706 &  &  0.708 & \textbf{0.725} & 0.708 & 0.739 &  & 0.691 & 0.721 & 0.704  & 0.699 &  &  \textbf{0.724} & 0.700 & \textbf{0.712} & \textbf{0.740} \\
& \textbf{BERT-S\bm{$\uparrow$}}     &  & 0.973 &  & 0.969     & 0.975 & \textbf{0.975} & 0.975 &  & 0.964 & 0.969 & 0.969  &0.968  &  & \textbf{0.975}  & 0.971 & \textbf{0.976} & \textbf{0.976} \\
\bottomrule
\end{tabular}%
}
\end{table*}

\begin{figure*}[t]
    \centering
    \subfigure[Loss with $\epsilon = 0.05 $ to $\epsilon =5$.]{\includegraphics[width=0.32\linewidth]{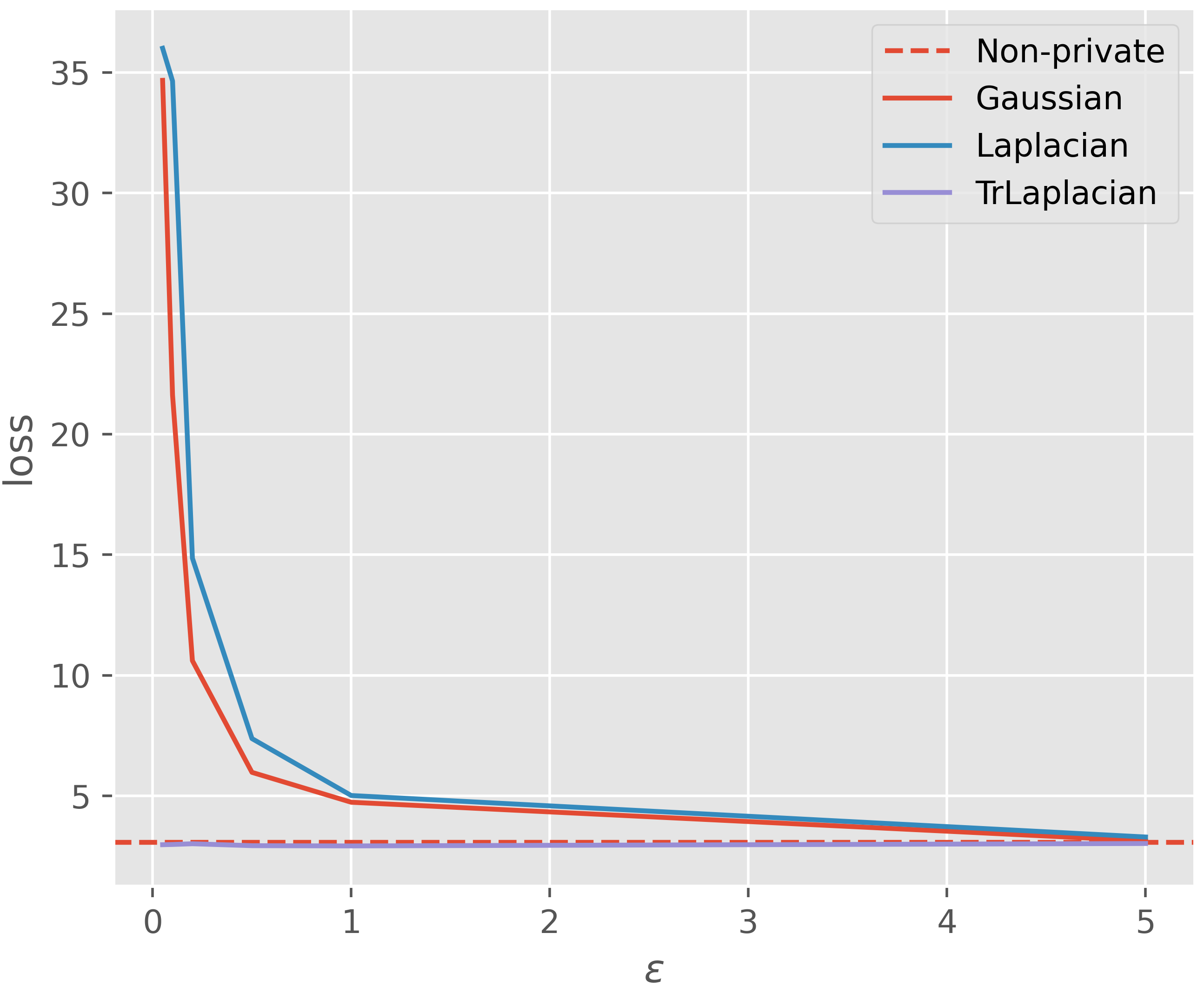}} 
    \subfigure[Rouge1 with $\epsilon = 0.05 $ to $\epsilon =5$.]{\includegraphics[width=0.32\linewidth]{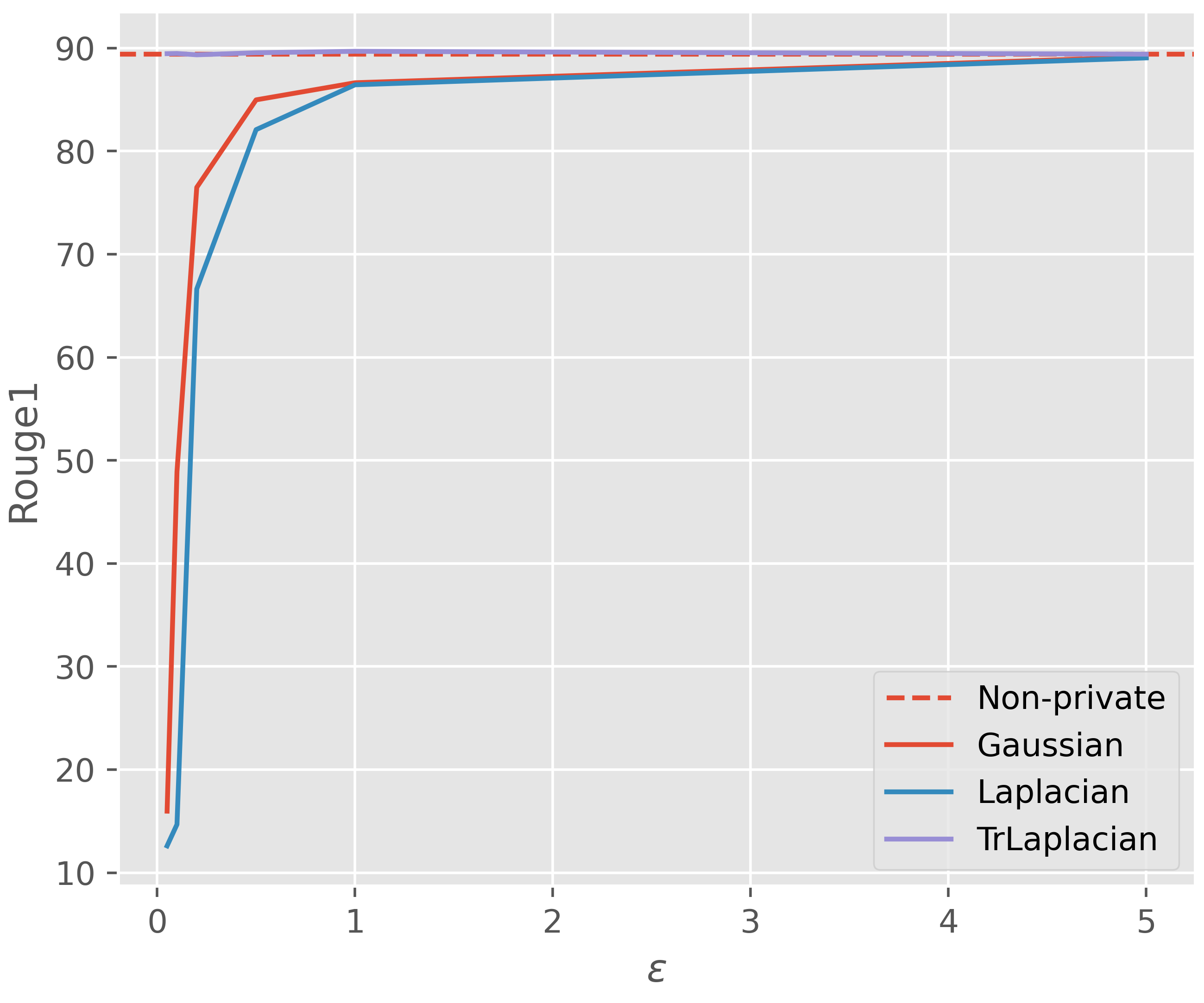}}
    \subfigure[BERT Score with $\epsilon = 0.05 $ to $\epsilon =1$.]{\includegraphics[width=0.32\linewidth]{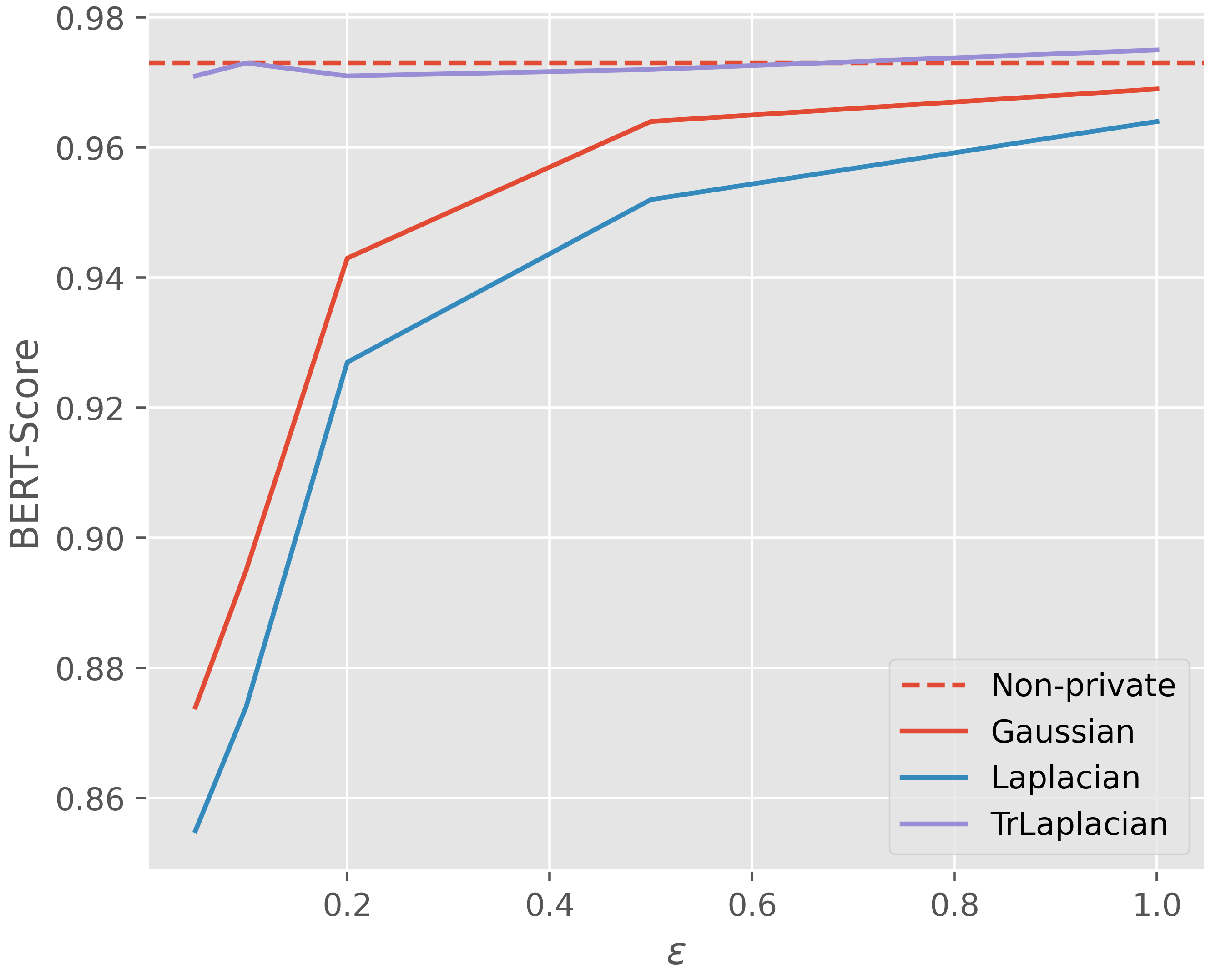}}
    \subfigure[Loss with $\epsilon = 0.05 $ to $\epsilon =5$.]{\includegraphics[width=0.32\linewidth]{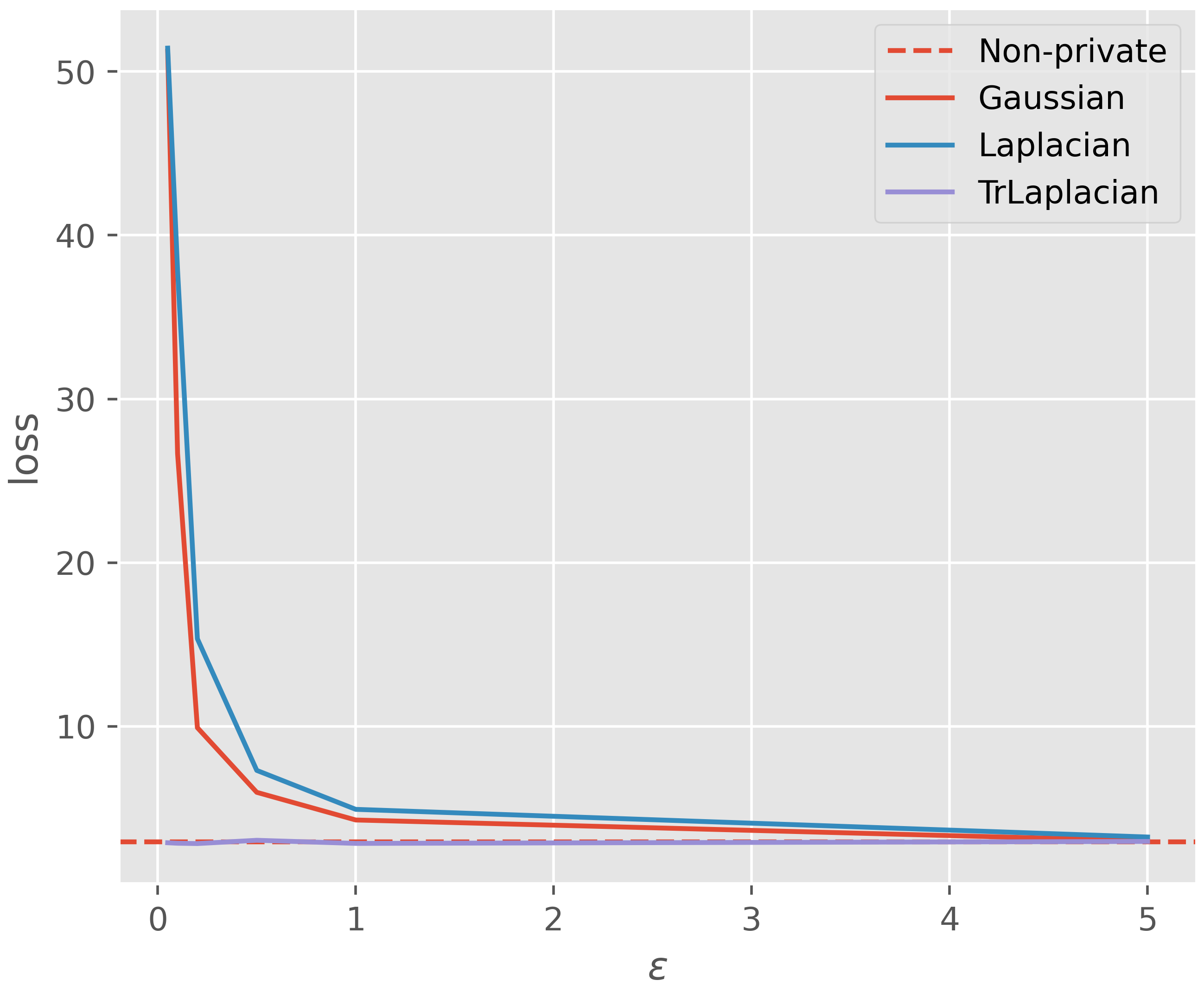}} 
    \subfigure[Rouge1 with $\epsilon = 0.05 $ to $\epsilon =5$.]{\includegraphics[width=0.32\linewidth]{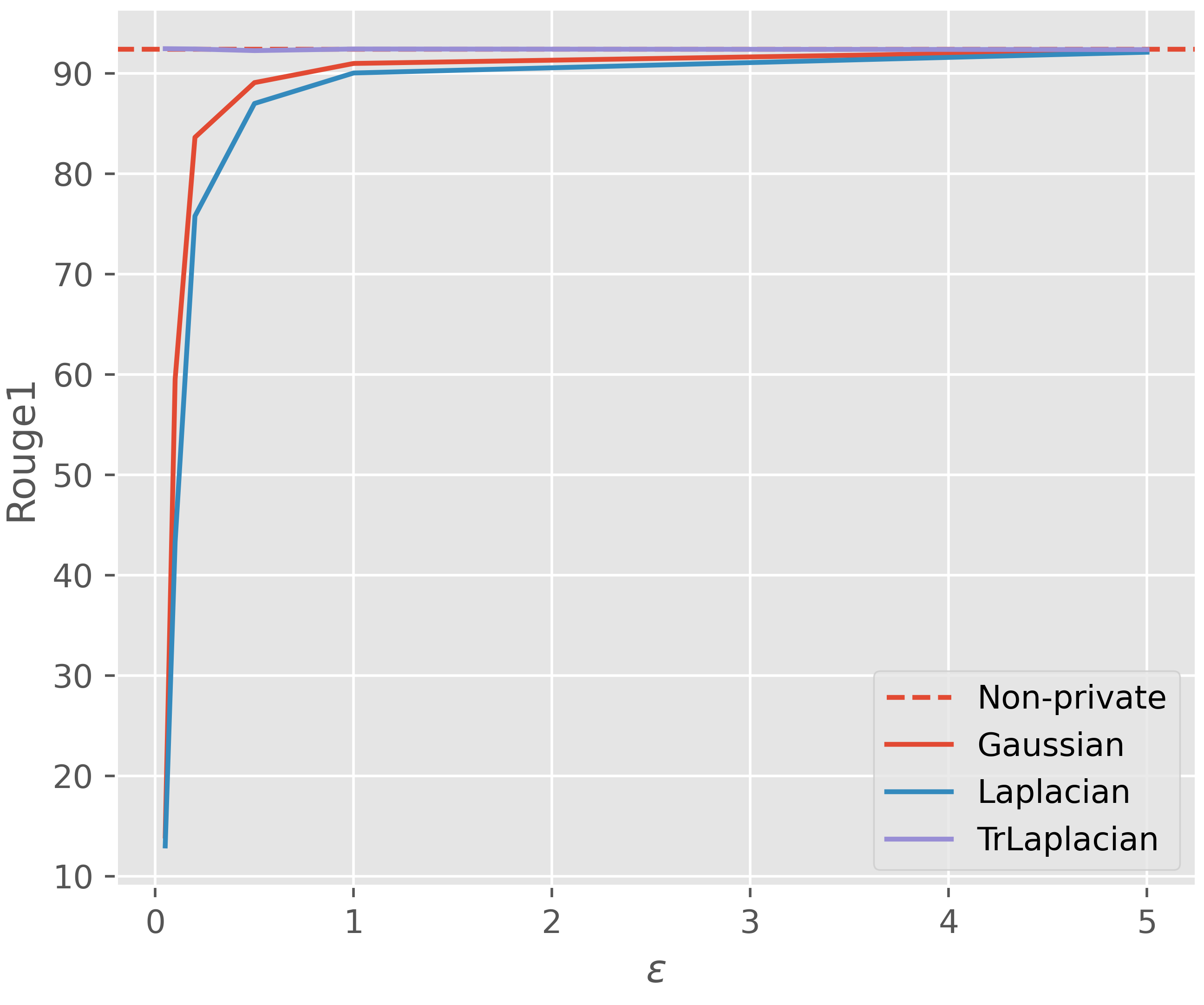}}
    \subfigure[BERT Score with $\epsilon = 0.05 $ to $\epsilon =1$.]{\includegraphics[width=0.32\linewidth]{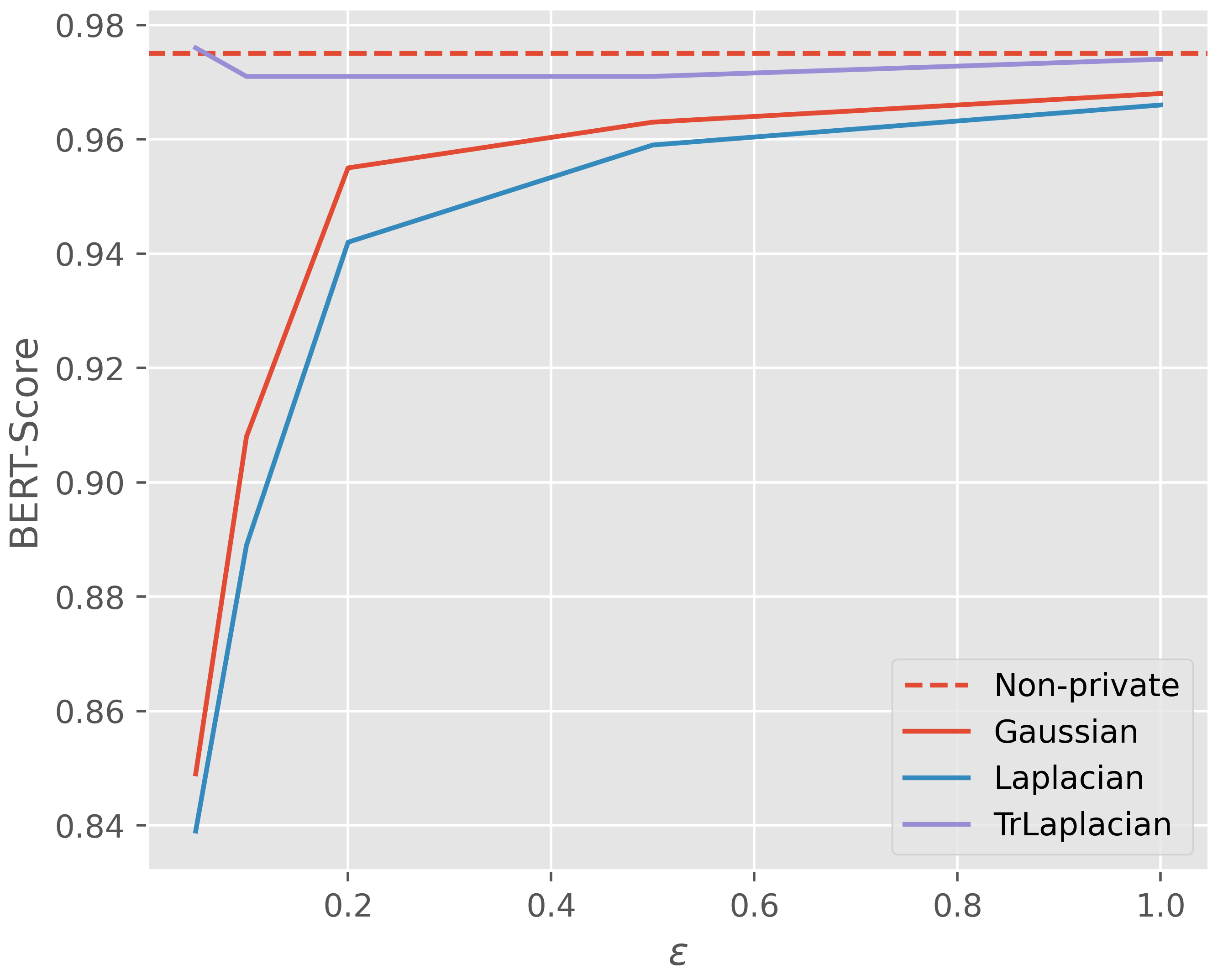}}
    \caption{ {\bf Privacy-Utility Test}. Curves of 
Loss, Rouge1 and BERTScore with different privacy budget $\epsilon$ for Yelp (Upper) and Yahoo (Lower) datasets.\label{fig:loss}}
\end{figure*}
\paragraph{Baseline.}
For DP text re-write, although \citet{krishna-etal-2021-adept} use the Laplacian mechanism to the sentence level DP instead of word level as in Definition \ref{def:1}. However, as \citet{habernal-2021-differential} mentions, the approach in \citet{krishna-etal-2021-adept} is not DP. Thus, here we will not compare with their method, and we will use the Laplacian and Gaussian mechanisms for the clipped embedding as baseline methods. For private fine-tuning, as we mentioned previously, all the previous methods only focus on metric DP instead of the original DP in Definition \ref{def:1}. Thus, our method is incomparable with theirs, and we will still use Laplacian and Gaussian mechanisms as baselines. 

For utility test, we compare our method with Gaussian and Laplacian mechanism for the original DP notation, as well as Calibrated Multivariate Perturbations (CMP) \citep{feyisetan2020privacy}, Mahalanobis Mechanism \citep{xu2020mahala} and Private Text Embedding (PTE) \citep{feyisetan2021private}, which are benchmark metric DP-based methods and have better utility than the original DP-based ones.

\paragraph{Evaluation Metrics.}
We use the loss of cross-entropy to measure the performance of language models. Specifically, cross-entropy is mainly used to determine how similar the actual output is to the expected output. Smaller model loss indicates less noise added to perturb the text. Additionally, we will use Rouge1 and BLEU scores. Rouge1 \citep{lin-2004-rouge} calculates recall using standard results and the number of 1-grams co-occurring in the auto-generated text. Similarly, BLEU \citep{papineni-etal-2002-bleu} measures the similarity between standard results and automatically generated text. Rouge1 measures word-level accuracy, while BLEU measures sentence fluency. Moreover, we use BERTScore \citep{Zhang2020BERTScore} to measure the semantic similarity of the perturbed sentence with the original one. To measure the privacy-preserving ability, we use the percentage of $N_w$ \citep{feyisetan2020privacy}, which is the number of words that are not replaced. Thus, under the same privacy budget, larger $N_w$ will be better (we want to change fewer words for accuracy).

\begin{figure*}[t]
    \centering
    \subfigure[SST-2, fastText (0.9013)]{\includegraphics[width=0.32\linewidth]{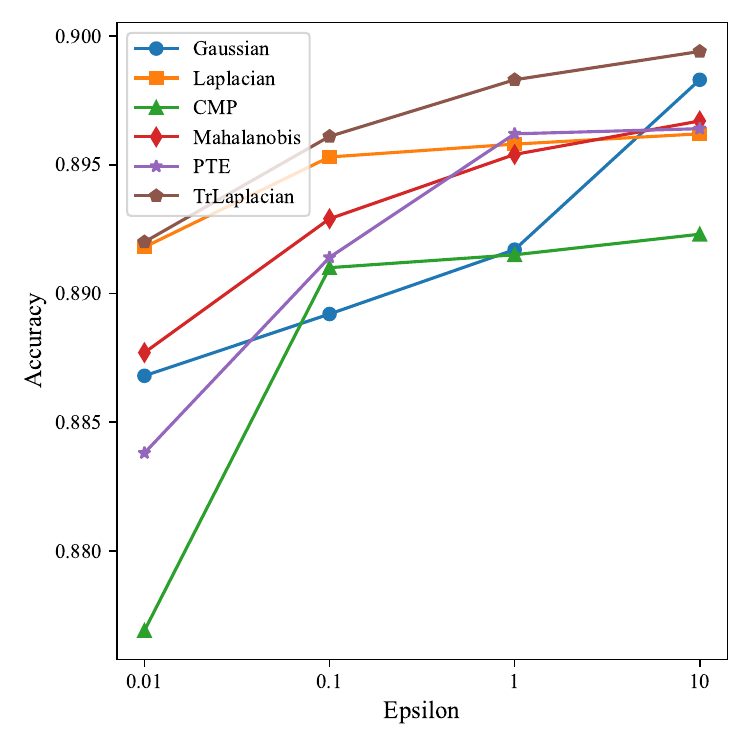}} 
    \subfigure[SST-2, GloVe (0.8997)]{\includegraphics[width=0.32\linewidth]{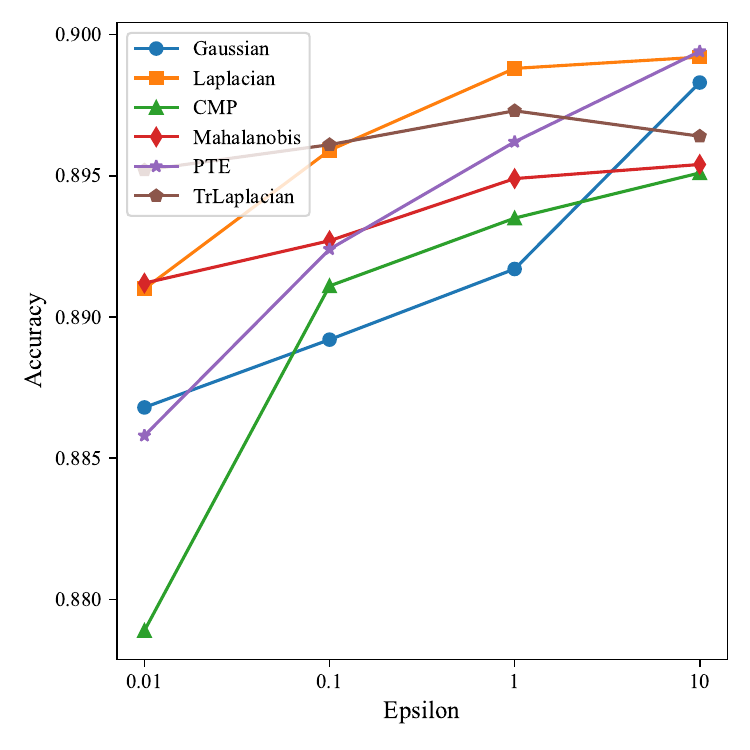}}
    \subfigure[SST-2, Random (0.5607)]{\includegraphics[width=0.32\linewidth]{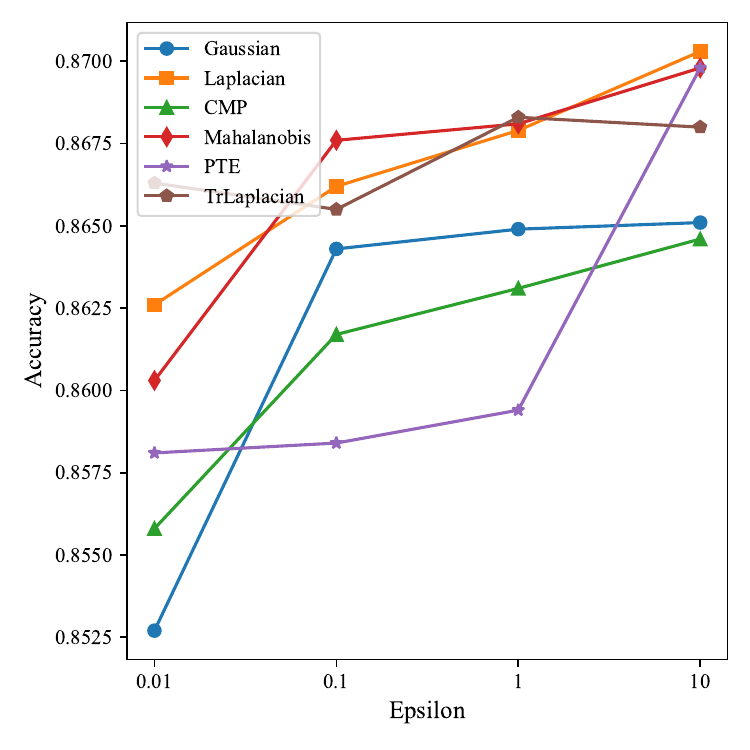}}
    \subfigure[AG News, fastText (0.9138)]{\includegraphics[width=0.32\linewidth]{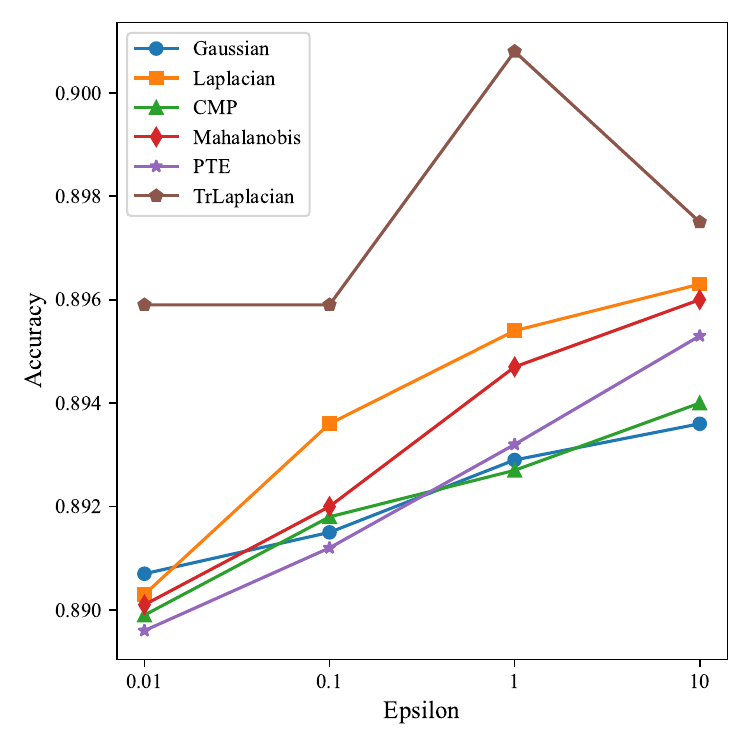}} 
    \subfigure[AG News, GloVe (0.9161)]{\includegraphics[width=0.32\linewidth]{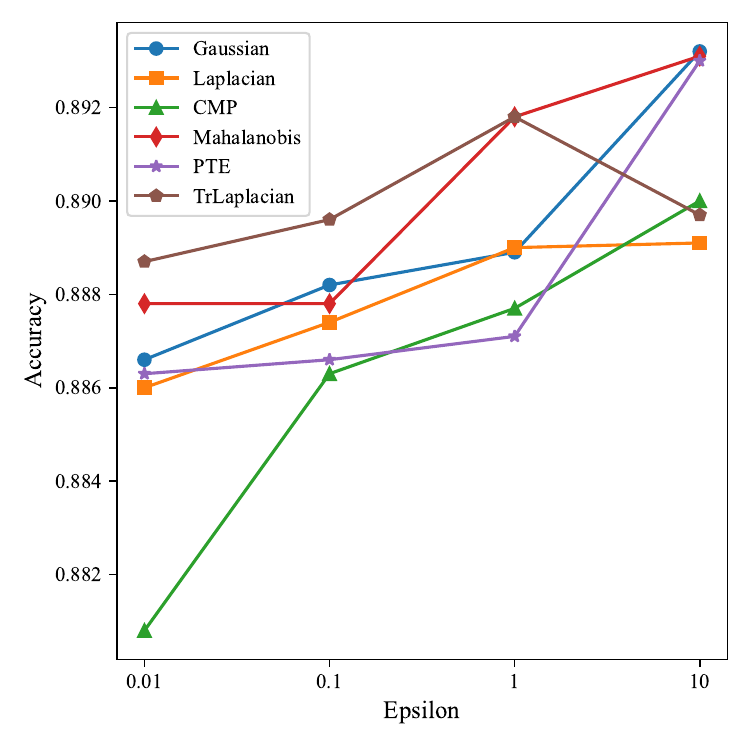}}
    \subfigure[AG News, Random (0.2462)]{\includegraphics[width=0.32\linewidth]{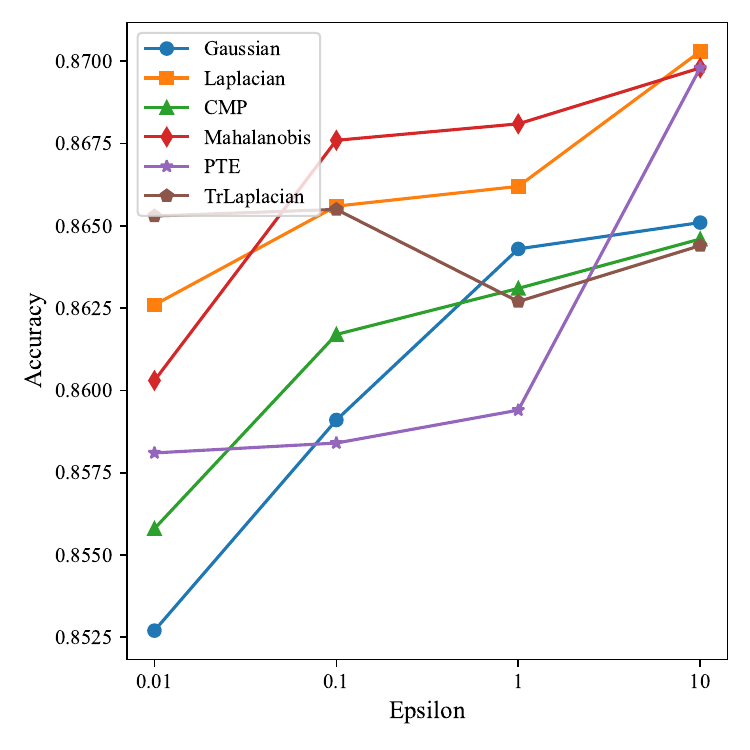}}
    \caption{Classification accuracy for all experimental settings. Each set of data is the average result of five experiments. We have included the baseline accuracy in parentheses in the subtitle of each subfigure. \label{fig:utility}}
\end{figure*}
\paragraph{Implementation Details.}
As an embedding can be considered as an initialization of the model, here we will consider three different initialization: Random embedding \citep{DBLP:conf/iclr/WietingK19}, GloVe \citep{DBLP:conf/emnlp/PenningtonSM14} and fastText \citep{DBLP:journals/tacl/BojanowskiGJM17}. We conduct experiments on these embeddings and the subsequent fine-tuning in the DP model via our mechanism. Each pre-trained word embedding is a 300-dimensional vector, and the size of considered vocabulary is $10^4$. For privacy budget, we set $\delta=\frac{1}{4^d}$, and we consider both the high privacy regime where $\epsilon\in \{0.05, 0.1, 0.2, 0.5\} $ and the low privacy regime $\epsilon\in \{1, 5, 10, 20\}$. For large $\epsilon$ we will use our previous dummy dimension trick ($d=500$ for $\epsilon = 10$ and $d = 1700$ for $\epsilon = 20$).

\subsection{Privacy Experiment on Embedding}
We first show the results on private embedding. Specifically, we use GloVe or fastText for the initialization, and then use three different private embedding mechanisms with different privacy budgets. Noted that large $\epsilon > 10$ is meaningless for privacy, we concentrate more on a small privacy budget in the main context. 

Fig. \ref{fig:exp} and \ref{fig:exp2} show the text after projecting the clipped and perturbed embedding back to the word domain in step 4 of Algorithm \ref{alg:1} for different mechanisms  when $\epsilon=0.1$. We can see our method (TrLaplace) outperforms the other two methods from both privacy and semantic perspectives, while the Gaussian mechanism fails to obfuscate the time, 
and the Laplacian mechanism totally replaces the time by another word, which destroys the structure of the sentence.  

Tab. \ref{tb:1} and Tab. \ref{tb:3} are the results on different metrics regarding private embedding with Glove initialization and Tab. \ref{tb:6} is with fastText initialization. We also  present the detailed trends w.r.t $\epsilon$ for three mechanisms in Fig. \ref{fig:loss}. 
When $\epsilon < 1$, from Tab. \ref{tb:1} 
we can see that for both Yahoo and Yelp, the loss of Gaussian and Laplacian mechanisms will be catastrophically large while our mechanism has a much smaller loss. From  Tab. \ref{tb:3} we can see we have almost the same phenomenon when in the low privacy regime.
Moreover, for Rouge1, Trlaplacian also surpasses the other two mechanisms on both datasets, indicating that our mechanism consistently demonstrates superiority in lexical/syntactic aspects.
For BLEU, the gap between all three mechanisms to the non-private case becomes small for both two datasets. But our method still has a slight advantage compared with the other two.

\begin{figure}[!ht]
    \centering
    \includegraphics[width=0.75\linewidth]{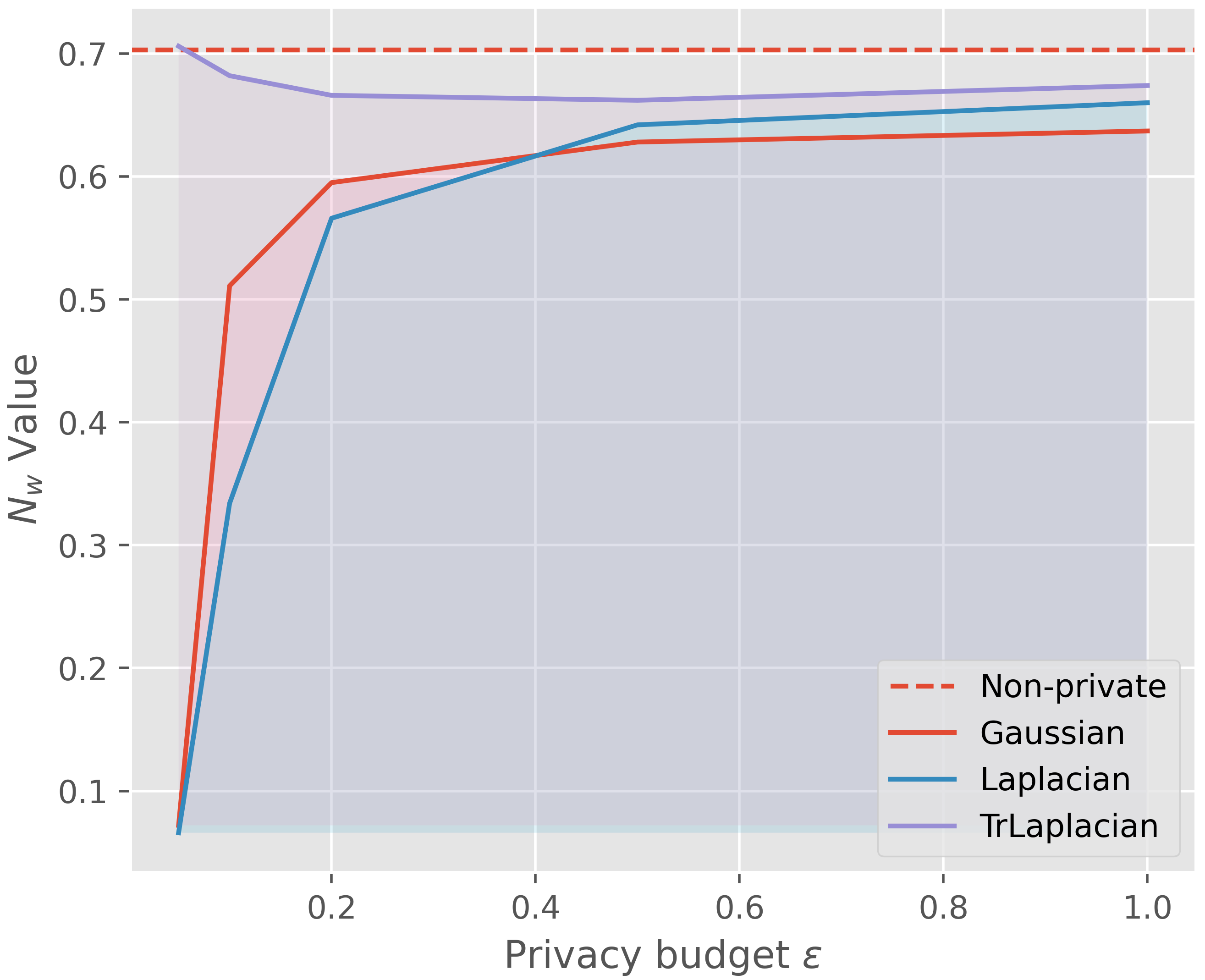}
    \caption{Privacy Test. Curves of  $N_w$ value w.r.t. privacy budget $\epsilon$  for Yahoo dataset.
    \label{fig:nw2}}
    \vspace{-5pt}
\end{figure}
For $N_w$ value, we can see in Fig. \ref{fig:nw} and Fig. \ref{fig:nw2}, our mechanism outperforms the other two mechanisms by changing less percentage of words to achieve the same privacy level, which indicates our method can exactly find sensitive words without hurting other words, thus keeps semantic properties.
For BERTScore, our mechanism is almost the same as the non-private case, while there is a larger gap for others. It is notable that, in almost all experiments our mechanism is the best, and the Gaussian mechanism is better than the Laplacian mechanism, which matches our theorem. However, it becomes less obvious when $\epsilon$ is large. The main reason is that when $\epsilon$ is enough large the noise will be sufficiently small and becomes nearly negligible, which can also be supported by the proof of Theorem \ref{thm:2}. For evaluation metrics, our mechanism may even be better than the non-private case, this may be due to small noise that could improve generalization, which is similar to adversarial training.  Moreover. in real-world scenarios, the privacy budget must be very small as we just want to keep the word level of privacy. This is because, in more realistic scenarios, when we want to protect the whole sentence, the total privacy budget required to privatize an entire sequence may grow linearly with its length (due to the composition theorem). Thus, the budget for each word should be extremely small. This has also been mentioned in some previous work~\cite{mattern-etal-2022-limits}.

\subsection{Utility of Private Fine-tuning}
We present the classification accuracy results for private fine-tuning across various embeddings and privacy levels in Fig.  \ref{fig:utility}. While it is acknowledged that utility will be affected by the size of the budget, with a smaller budget potentially leading to lower utility, our experimental results show that our proposed method maintains relatively stable utility across different budget choices. Specifically, even in the high privacy regime ($\epsilon=0.01$), our approach only incurs a slight decrease in utility compared to the non-private scenario. Similar capabilities of other methods can be observed in the experimental results of \citet{meisenbacher2024comparative}.

We can also observe the effect of different pretrained embeddings. It is evident that when using GloVe and fastText pretrained embeddings, all methods achieve accuracy close to the baseline. However, when using random embeddings, the baseline accuracy is very low (0.5607 on SST-2, 0.2462 on AG News), but all methods significantly improve the accuracy of downstream sentiment analysis and topic classification tasks after training, with accuracy reaching up to about 0.87 when epsilon=10. This indicates that all methods are effective after training. Specifically, our method, Truncated Laplacian, maintains an accuracy of around 0.865 when trained on random embeddings, demonstrating that it maintains privacy while preserving the quality of embeddings, thereby offering excellent performance.




\section{Conclusions}
We introduce a novel method called the high dimensional truncated Laplacian mechanism for private embedding, which extends the one-dimensional case to the high-dimensional case. Theoretical analysis demonstrates that our method exhibits lower variance compared to existing private word embedding techniques. Experiments show that even in the high privacy regime, our approach incurs only a minimal loss in utility compared to the non-private case, which maintains privacy while preserving the quality of embeddings for promising performance.

\section{Limitations}
First, the word level DP has the disadvantages of length constraints and linear growth of privacy budget \citep{mattern-etal-2022-limits}. However, such limitations are rooted in the definition of DP instead of our mechanism.  Secondly, to ensure DP guarantees, in this paper, our mechanism involves clipping embedding vectors and adding calibrated noises, which inevitably introduce errors to the outputs of the task at hand. And these errors may affect different groups of individuals differently and may cause unfairness issues. However, we still need to mention that such unfairness issues are mainly due to the definition of DP rather than our method, as DP machine learning algorithms will always have a disparate impact on model accuracy \citep{DBLP:conf/nips/BagdasaryanPS19}. Despite some limitations, word-level DP still offers unique advantages and potential applications~\citep{hu-etal-2024-differentially}, and brings value to the DP-NLP community.

\section{Ethics Review}
This paper presents work whose goal is to advance the field of NLP. There are many potential societal consequences of our work, none which we feel must be specifically highlighted here.

\section*{Acknowledgments}
Di Wang and Lijie Hu are supported in part by the funding BAS/1/1689-01-01, URF/1/4663-01-01,  REI/1/5232-01-01,  REI/1/5332-01-01, and URF/1/5508-01-01  from KAUST, and funding from KAUST - Center of Excellence for Generative AI, under award number 5940. Ivan Habernal is supported by the Research Center Trustworthy Data Science and Security (\url{https://rc-trust.ai}), one of the Research Alliance centers within the UA-Ruhr (\url{https://uaruhr.de}).

\bibliography{anthology,custom}

\appendix

\section{More Details and Experiments\label{dataset}}
\label{Appendix1: datastatistics}

\paragraph{Dataset.}
The statistics of dataset are shown in table \ref{tb:7}.
\begin{table}[htbp]
    \centering
    \small
    \caption{Dataset statistics used in this work. \label{tb:7}}
    \begin{tabular}{lccc}
        \toprule
        Dataset & Avg. Length & Train Size & Test Size \\
         & (tokens) & (neg/pos) & (neg/pos) \\
        \midrule
        Yahoo & 181 & 8539/8673 & 2174/2189 \\
        Yelp & 19 & 3610/3310 & 909/912 \\
        SST-2 & 10 & 24214/30362 & 5994/7651 \\
        AG News & 40 & 20000 & 4000 \\
        \bottomrule
    \end{tabular}
\vspace{-10pt}
\end{table}

\begin{table}[htbp]
\centering
\caption{\textbf{Time Cost.} Comparison of the time cost of each epoch (seconds) under GloVe Embedding initialization for  the non-private case and three mechanisms (Gaussian, Laplacian and TrLaplacian), the privacy budget ranges from 0.05 to 20. \label{tb:8}}
\resizebox{0.48\textwidth}{!}{%
\begin{tabular}{cclcccclcccc}
\toprule
& &  & \multicolumn{4}{c}{\bm{$\epsilon<1$}}           &  & \multicolumn{4}{c}{\bm{$\epsilon \geq 1$}}         
\\ \cline{4-7} \cline{9-12}  
\multicolumn{2}{c}{\textbf{Privacy budget $\bm{\epsilon}$}}     &  & 0.05   & 0.1 & 0.2 & 0.5 &    & 1   & 5 & 10 & 20 \\ 
\midrule
\multirow{4}{*}{\textbf{Yahoo}}& \textbf{Non-private}   &  & \multicolumn{9}{c}{111}\\
& \textbf{Gaussian}       &  & 111 & 113  &  111  &  111 & &       111   & 111   & 111 & 111     \\
& \textbf{Laplacian}       &  & 111 &  113 & 111   & 111 &  &       111    &  111  & 111 & 111   \\
\rowcolor{gray!30}
& \textbf{TrLaplacian}       &  & 123 & 123  & 123   & 123 &  &         123  &  123  & 123 & 123   \\ 
\midrule
\multirow{4}{*}{\textbf{Yelp}}& \textbf{Non-private}  & & \multicolumn{9}{c}{111}\\
& \textbf{Gaussian}       &  & 38 &  37 &  38  & 38 &  &      37     & 37   & 37 &  37    \\
& \textbf{Laplacian}       &  & 38 & 37  &  37  & 37 &  &       37    &  37  & 37 & 37   \\
\rowcolor{gray!30}
& \textbf{TrLaplacian}       &  & 46 & 41 & 46   & 42 &  &         42  &   42 & 42 & 42   \\ 
\bottomrule
\end{tabular}%
}
\end{table}

\paragraph{Implementation Details.} 
Models in this paper are implemented based on the PyTorch \footnote{https://pytorch.org/} and TensorFlow \footnote{https://www.tensorflow.org/} with their libraries. Experiments are conducted on NVIDIA GeForce RTX 3090 GPUs.  To implement our mechanism, we use the acceptance-rejection sampling method~\cite{accept-reject} to sample a point from the high dimensional truncated Laplace distribution from the Laplace distribution, only by rejecting the samples outside the interval.

For text re-write, we use the auto-encoder model. The embedding is initialized with the 300-dimensional pre-trained Random, GloVe, and fastText word embedding. We use one-layer BiLSTM with dropout for the encoder, and using setup: dropout rate 0.5, Adam \citep{Kingma2015AdamAM} with an initial learning rate of 0.01 and betas (0.5, 0.999), batch size 1024, and number of training epochs 100.
For the downstream classification task over the AG News and SST-2 dataset, we use Adam with an initial learning rate of 0.0001, a dropout rate of 0.2, and a batch size of 256. We set the maximum number of epochs to be 50. 



\begin{figure*}[!htbp]
\centering
\tikzset{
    >=stealth',
    punkt/.style={
          very thick,
          rectangle split,
          rectangle split parts=2,
          inner ysep=1.2mm,
          inner xsep=1.5mm,
          rounded corners,
          draw=black, thick,
          text width=9.2cm},
    punx/.style={
          rectangle,
          rounded corners,
          align=left,
          text width=0.55cm,
          text depth=0.01cm,
          draw=white},
  header/.style = {%
    inner ysep = +1.5em,
    append after command = {
      \pgfextra{\let\TikZlastnode\tikzlastnode}
      node [header node] (header-\TikZlastnode) at (\TikZlastnode.north) {#1}
      node [span = (\TikZlastnode)(header-\TikZlastnode)]
        at (fit bounding box) (h-\TikZlastnode) {}
    }
  },
}

\begin{tikzpicture}[scale=1.0]
\node[punkt](Vanilla){
{\textbf{Comparison Semantic Problem of Private Embedding} \quad }
\nodepart{second}
{\centering \small \textbf{Original}:~\,
\colorword{pink!20}{black}{do}
\colorword{pink!100}{black}{not}
\colorword{pink!20}{black}{come here! food}
\colorword{pink!100}{black}{poisoning alert!} 
($\rightarrow$\textcolor{blue}{Neg.})
}\\
{\centering \small \textbf{Trlaplace}:\, 
\colorword{pink!20}{black}{do}
\colorword{pink!100}{black}{not}
\colorword{pink!20}{black}{come here! food}
\colorword{pink!100}{black}{poisoning alert!}
($\rightarrow$\textcolor{blue}{Neg.})
}\\
{\centering \small \textbf{Laplace}:~\,
\colorword{pink!20}{black}{this place is} \colorword{pink!100}{black}{awesome!} 
\colorword{pink!100}{black}{love} 
\colorword{pink!20}{black}{this place!}
($\rightarrow$\textcolor{red}{Pos.})
}\\
{\centering \small \textbf{Gaussian}:\,
\colorword{pink!20}{black}{do}
\colorword{pink!100}{black}{not}
\colorword{pink!20}{black}{go here! food}
\colorword{pink!100}{black}{glorious <unk>!}
($\rightarrow$\textcolor{red}{Pos.})
}
};
\end{tikzpicture}
\caption{Another example of text re-write with different mechanisms with $\epsilon=0.1$. The Gaussian and Laplacian mechanisms destroyed the semantic properties of the original sentence. \label{fig:exp2}}
\end{figure*}
\section{Omitted Proofs}
\begin{Proof}[Proof of Theorem \ref{thm:1}] The proof is motivated by \cite{DBLP:journals/corr/abs-2106-07094}. 
Consider a pair of tokens $w, w'$. Let perturbed encoder1 $r_1 = \operatorname{CLIPEmb}(w)+\eta_1$, also let $r_2 = \operatorname{CLIPEmb}(w')+\eta_2= \operatorname{CLIPEmb}(w) +\boldsymbol{\Delta}_{\boldsymbol{s}}+\eta_2$, where $\left\|\boldsymbol{\Delta}_{\boldsymbol{s}}\right\|_{1} \leq \Delta_{1}$ and $\left\|\boldsymbol{\Delta}_{\boldsymbol{s}}\right\|_{\infty} \leq \Delta_{\infty}$ which are due to the clipping operation.

Let us denote the set of possible values of $r_k$ by $\mathcal{S}_{k}$ for $k={1,2}$. 

Define $\mathcal{U}=[-C-A,C+A]^d$. Note that for any subset $\mathcal{V} \subseteq \mathcal{U}-(\mathcal{S}_{1} \cup \mathcal{S}_{2})$, $\mathbb{P}\left(r_1 \in \mathcal{V}\right)  =\mathbb{P}\left(r_2 \in \mathcal{V}\right)=0$, hence $(\epsilon, \delta)$-DP is satisfied for this part. We need to ensure $(\epsilon, \delta)$-DP is satisfied for all elements in $\mathcal{S}_{1} \cup \mathcal{S}_{2}$ too.

First, consider an element $s \in \mathcal{S}_{1} \cap \mathcal{S}_{2}$. Then:

\begin{equation*}
\begin{split}
f \left(r_1=s \right) &= f  \left(\eta_1=s-\operatorname{CLIPEmb}(\mathbf{s}) \right)
\end{split}
\end{equation*}

Similarly:

\begin{equation*}
\begin{split}
f\left(r_2=s \right)
&= f\left(\eta_2=s-
\operatorname{CLIPEmb}(\mathbf{s})  -\boldsymbol{\Delta}_{\boldsymbol{s}} \right)
\end{split}    
\end{equation*}

Using the above equations:
\begin{equation*}
\begin{split}
& \exp \left(-\alpha \Delta_{1}\right) \leq \exp \left(-\alpha\left\|\Delta_{s}\right\|_{1}\right) \\
& \leq \frac{f\left(r_1=s \right)}{f\left(r_2=s \right)} \leq \exp \left(\alpha\left\|\Delta_{s}\right\|_{1}\right) \leq \exp \left(\alpha \Delta_{1}\right)    
\end{split}    
\end{equation*}

From the above equation, setting setting $\alpha=\epsilon / \Delta_{1}$ ensures pure $\epsilon$-DP for all $s \in \mathcal{S}_{1} \cap \mathcal{S}_{2}$. With this, it follows that for any $\mathcal{V} \subseteq \mathcal{S}_{1} \cap \mathcal{S}_{2}$:
\begin{equation*}
e^{-\epsilon} \mathbb{P}\left(r_2 \in \mathcal{V} \right) \leq \mathbb{P}\left(r_1 \in \mathcal{V} \right) \leq e^{\epsilon} \mathbb{P}\left(r_2 \in \mathcal{V} \right).   
\end{equation*}
by setting $\alpha=\epsilon / \Delta_{1}$.

Now consider an element $s \in \mathcal{S}_{2}-\mathcal{S}_{1}$. Clearly, $f\left(r_1=s\right)=0$. Also:
\begin{equation*}
\max _{s \in \mathcal{S}_{2}-\mathcal{S}_{1}} f\left(r_2=s\right) \leq \frac{1}{B}.
\end{equation*}

But notice that $\text{volume}(\mathcal{S}_{2}-\mathcal{S}_{1}) \leq \Delta_{\infty}^{d}$. This follows from the fact that for every coordinate, there are at most $\Delta_{\infty}$ levels that can be attained by $r_2$ but not by $r_1$. Thus, for any 
$\mathcal{T} \subseteq \mathcal{S}_{2}-\mathcal{S}_{1}$, we have
\begin{equation*}
\mathbb{P}\left(r_1 \in \mathcal{T}\right)=0 \text { and } \mathbb{P}\left(r_2 \in \mathcal{T}\right) \leq\left(\frac{\Delta_{\infty}}{B}\right)^{d}
\end{equation*}
Similarly, for any $\mathcal{T} \subseteq \mathcal{S}_{1}-\mathcal{S}_{2}$, we have
\begin{equation*}
\mathbb{P}\left(r_2 \in \mathcal{T}\right)=0 \text { and } \mathbb{P}\left(r_1\in \mathcal{T}\right) \leq\left(\frac{\Delta_{\infty}}{B}\right)^{d}.
\end{equation*}

Now, let us now consider some general $\mathcal{T} \subseteq \mathcal{S}_{1} \cup \mathcal{S}_{2}$. Let $\mathcal{T}_{0}=\mathcal{T} \cap\left(\mathcal{S}_{1} \cup \mathcal{S}_{2}\right), \mathcal{T}_{1}=\mathcal{T} \cap\left(\mathcal{S}_{1}-\mathcal{S}_{2}\right)$ and $\mathcal{T}_{2}=\mathcal{T} \cap\left(\mathcal{S}_{2}-\mathcal{S}_{1}\right)$. It is easy to see that $\mathcal{T}=\mathcal{T}_{0} \cup \mathcal{T}_{1} \cup \mathcal{T}_{2}$ and that $\mathcal{T}_{0}, \mathcal{T}_{1}$ and $\mathcal{T}_{2}$ are pairwise-disjoint. Then:
\begin{equation}
\begin{split}
\mathbb{P}\left(r_1 \in \mathcal{T}\right) &=\mathbb{P}\left(r_1 \in \mathcal{T}_{0}\right)+\mathbb{P}\left(r_1 \in \mathcal{T}_{1}\right)\\
&+\mathbb{P}\left(r_1 \in \mathcal{T}_{2}\right)\\
& \leq e^{\epsilon} \mathbb{P}\left(r_2 \in \mathcal{T}_{0}\right)+\left(\frac{\Delta_{\infty}}{B}\right)^{d}+0 \\
& \leq e^{\epsilon} \mathbb{P}\left(r_2 \in \mathcal{T}\right)+\left(\frac{\Delta_{\infty}}{B}\right)^{d}.
\end{split}
\end{equation}
Thus, we can set $\delta=(\frac{\Delta_{\infty}}{B})^d$. Obviously, this result is only useful if $B>\Delta_{\infty}$. 

For each coordinate
\begin{equation*}
\begin{split}
& \int_{x \in \mathbb{R}} f_{\mathrm{TLap}}(x) d x=\int_{0}^{A} 2 \frac{1}{B} e^{-\alpha|x|} dx \\
&=\frac{2}{B\alpha} \left(1-e^{-\alpha A}\right)=1        
\end{split}
\end{equation*}

We can solve $B=\frac{2(1-e^{-\alpha  A})}{\alpha}$. Thus, take $B=\frac{\Delta_\infty}{\delta^\frac{1}{d}}$, we can see $A=-\frac{1}{\alpha} \log (1-\frac{\alpha\Delta_\infty  }{2\delta^\frac{1}{\delta}} )=-\frac{\Delta_1}{\epsilon}\log (1-\frac{\epsilon}{2\sqrt{d} \delta^\frac{1}{\delta}})$.
\end{Proof}
\begin{Proof}[Proof of Theorem \ref{thm:2}]
We first show the variance of our mechanism $\mathcal{A}$ is bounded by $2\frac{d\Delta_1^2}{\epsilon^2}$. 

We can easily see that the variance is $\mathbb{E}\|\mathcal{A}(w)-w\|_2^2=dV$ with $V =\int_{x \in \mathbb{R}} f_{\mathrm{TLap}}(x)|x|^2 d x$, so 

\begin{equation}
\begin{split}
& \int x^{2} f(x) d x \\
=& 2 \frac{1}{B} \int_{0}^{A} e^{-\alpha x} x^{2} d x \\
=& 2 \frac{1}{B}\int_{0}^{A}-\frac{1}{\alpha} x^{2} d\left(e^{-\alpha x}\right) \\
=& 2 \frac{1}{B}(-\frac{1}{\alpha}) A^{2} e^{-\alpha A}+2 \frac{1}{B} \int_{0}^{A} \frac{1}{\alpha} e^{-\alpha x} 2 x d x    
\end{split}
\end{equation}
and 
\begin{equation}
\begin{split}
& \int_{0}^{A} \frac{1}{\alpha} e^{-\alpha x} 2 x d x \\
=&-\int_{0}^{A} \frac{1}{\alpha^{2}} \cdot 2 x d\left(e^{-\alpha x}\right) \\
=&-\frac{1}{\alpha^{2}} 2 A e^{-\alpha A}+\int_{0}^{A} \frac{2}{\alpha^{2}} e^{-\alpha x} d x \\
=&-\frac{1}{\alpha^{2}} 2 A \cdot e^{-\alpha A}+\frac{2}{\alpha^3}\left(1-e^{-2\alpha A}\right)    
\end{split}
\end{equation}
Thus, we have
\begin{equation}\label{less}
\begin{split}
V&=-2 \frac{1}{\alpha} \frac{1}{B} A^{2} e^{-\alpha A}-4 \frac{1}{\alpha^{2}} \frac{1}{B} A e^{-\alpha A}\\
& +4 \frac{1}{\alpha^3} \frac{1}{B}\left(1-e^{-\alpha A}\right) \\
&=-2 \frac{1}{\alpha} \frac{1}{B} A e^{-\alpha A}(A+2 \frac{1}{\alpha})+2 \frac{\Delta_{1}^{2}}{\varepsilon^{2}} \\
& < 2 \frac{\Delta_{1}^{2}}{\varepsilon^{2}}. 
\end{split}
\end{equation}
Thus, in total we have $\mathbb{E}\|\mathcal{A}(w)-w\|_2^2\leq \frac{2d\Delta_1^2}{\epsilon^2}=\frac{8d^2C^2}{\epsilon^2}$.

Next for Laplacian mechanism in Theorem \ref{lap} we have 
$\mathbb{E}[\|\mathcal{A}_{lap}(w)-w\|_2^2]=\frac{2d\Delta_1^2}{\epsilon^2}$. Thus the variance of high dimensional truncated Laplacian is always lower than Laplacian. 

Similarly, the variance of Gaussian mechanism  in Theorem \ref{thm:2} is $\frac{8C^2d(\ln 1.25+\ln 1/\delta)}{\epsilon^2}$, we can easily see that our mechanism has lower variance when $\delta\leq \frac{1}{e^d}$. 
\end{Proof}

\end{document}